\documentclass[10pt,twocolumn]{article} 
\usepackage{primeARXIV}
\usepackage{times}
\usepackage{graphicx}
\usepackage{amssymb}
\usepackage{multirow}
\usepackage{algpseudocode}
\usepackage{algorithm}
\usepackage{array}
\usepackage{float}
\usepackage{arydshln}
\usepackage{soul}
\usepackage{placeins}
\usepackage{url,hyperref}
\usepackage{amsmath}
\usepackage{tikz}
\usepackage[format=plain,labelformat=simple,labelsep=period,font=small,compatibility=false]{caption}
\usepackage[font=footnotesize,skip=3pt,subrefformat=parens]{subcaption}

\usepackage{cleveref}
\crefname{section}{Sec.}{Secs.}
\Crefname{section}{Section}{Sections}
\crefname{figure}{Fig.}{Figs.}
\Crefname{figure}{Figure}{Figures}
\Crefname{table}{Table}{Tables}
\crefname{table}{Tab.}{Tabs.}

\newcommand\copyrighttext{%
  \footnotesize \textcopyright 2025 IEEE. Personal use of this material is permitted.
  Permission from IEEE must be obtained for all other uses, in any current or future
  media, including reprinting/republishing this material for advertising or promotional
  purposes, creating new collective works, for resale or redistribution to servers or
  lists, or reuse of any copyrighted component of this work in other works.}
\newcommand\copyrightnotice{%
\begin{tikzpicture}[remember picture,overlay]
\node[anchor=south,yshift=10pt] at (current page.south) 
  {\fbox{\parbox{\dimexpr\textwidth-\fboxsep-\fboxrule\relax}{\copyrighttext}}};
\end{tikzpicture}%
}
\newcommand{\mycomment}[1]{}
\newcommand{\toaddminor}{}
\newcommand{\toadd}{}
\newcommand{\addboxminor}{}
\newcommand{\addbox}{}

\begin{document}

\title{Variational Bayes image restoration with compressive autoencoders}

\author{Maud Biquard\\
ISAE-Supaero / CNES\\
31400 Toulouse, France\\
{\tt\small maud.biquard@isae-supaero.fr}
\and
Florence Genin, Christophe Latry\\
CNES\\
31400 Toulouse, France\\
{\tt\small firstname.lastname@cnes.fr}
\and
Marie Chabert\\
IRIT/INP-ENSEEIHT\\
31000 Toulouse, France\\
{\tt\small marie.chabert@toulouse-inp.fr}
\and
Thomas Oberlin\\
ISAE-Supaero\\
31400 Toulouse, France\\
{\tt\small thomas.oberlin@isae-supaero.fr}
}

\maketitle
\copyrightnotice
\thispagestyle{empty}

\begin{abstract}
Regularization of inverse problems is of paramount importance in computational imaging. The ability of neural networks to learn efficient image representations has been recently exploited to design powerful data-driven regularizers. While state-of-the-art plug-and-play (PnP) methods rely on an implicit regularization provided by neural denoisers, alternative Bayesian approaches consider Maximum A Posteriori (MAP) estimation in the latent space of a generative model, thus with an explicit regularization.
However, state-of-the-art deep generative models require a huge amount of training data compared to denoisers. Besides, their complexity hampers the optimization involved in latent MAP derivation. 
In this work, we first propose to use compressive autoencoders instead. These networks, which can be seen as variational autoencoders with a flexible latent prior, are smaller and easier to train than state-of-the-art generative models. As a second contribution, we introduce the Variational Bayes Latent Estimation (VBLE) algorithm, which performs latent estimation within the framework of variational inference. Thanks to a simple yet efficient parameterization of the variational posterior, VBLE allows for fast and easy (approximate) posterior sampling.
Experimental results on image datasets BSD and FFHQ demonstrate that VBLE reaches similar performance 
\toadd{as} state-of-the-art PnP 
methods, while being able to quantify uncertainties significantly faster than other existing posterior sampling techniques. \toadd{The code associated to this paper is available in \url{https://github.com/MaudBqrd/VBLE}.}
\end{abstract}

\textbf{Index terms---}
\textit{Image restoration, Inverse problems, Deep regularization, Posterior sampling}

\section{Introduction}
\label{sec:intro}

\noindent Image restoration tasks, such as deblurring, inpainting\toadd{,} or super-resolution, consist in recovering a clean image $x$ from its noisy measurement $y$, based on the forward model $y = Ax + w$, with
$A$ representing the degradation operator and $w$ an additive noise term. 
Within the Bayesian framework, this ill-posed inverse problem is typically solved by finding the Maximum A Posteriori (MAP) estimate, solution to the problem
\begin{equation}
    \max_x p_{X|Y} (x|y) \Leftrightarrow \min_x - \log p_{Y|X}(y|x) - \log p_X(x) \label{sol_inv_pb_var}
\end{equation}
where $\log p_{Y|X}(y|x)$ is the observation log-likelihood, which is quadratic for Gaussian noise,  
and $- \log p_X(x)$ acts as a regularization term, promoting solutions that are most compatible with the prior distribution $p_X(x)$.
Classical regularizations
include 
total variation \cite{Rudin1992}, Tikhonov regularization \cite{Tikhonov1963}, and sparsity-promoting penalties on well-chosen representations such as wavelet bases or dictionaries \cite{Tibshirani1996,Elad2010,Mallat2008}.

Deep learning has led to substantial performance gains in image restoration tasks. 
A first category of methods directly solves the inverse problem after a supervised end-to-end training on a dataset of original-degraded image pairs related by the forward model
\cite{Dong2014,Saharia2022}. 
Their performance 
\toadd{is} impressive but only for the specific inverse problem considered during training. For that reason, a second category of methods
aims at learning the regularization only, leveraging the forward model to solve the inverse problem. 
Besides offering better interpretability, they allow to solve a wide range of inverse problems with the same neural network.

Among them, plug-and-play (PnP) image restoration methods 
\cite{Venkatakrishnan2013}
yield excellent performance on a wide variety of image restoration tasks. 
These methods classically use splitting algorithms, such as ADMM \cite{Venkatakrishnan2013,Chan2016} or HQS \cite{Zhang2021,Hurault2022}, that separately handle the data term and the regularization term in the optimization problem (\ref{sol_inv_pb_var}).
Their key concept is to use Gaussian denoisers \cite{Kamilov2017}, and in particular deep denoisers \cite{Zhang2017a} in place of the proximal operator of the regularization. 
More recently, the introduction of generative denoisers, such as denoising diffusion models \cite{Ho2020,Nichol2021}, within the PnP framework, has demonstrated remarkable performance in solving inverse problems \toadd{\cite{Kawar2022, Chung2022, Zhu2023, Cao2024, Rout2024, Garber2024}}.

A more direct way to learn the regularization consists in 
estimating the data distribution
within generative models. This line of work, 
referred to as latent optimization in the following, seeks the inverse problem solution in the latent space of a generative model.
Specifically, given a generative model $G$ which has been trained on a dataset of ideal images
, the seminal work \cite{Bora2017} computes a MAP estimate in its latent space by solving the following optimization problem using gradient descent:
\begin{equation}
    \max_z p_{Z|Y} (z|y) \Leftrightarrow \min_z - \log p_{Y|Z}(y|z) - \log p_Z(z), \label{eq_bora}
\end{equation}
where $z$ is the latent variable to be optimized, $p_Z(z)$ is the generative latent prior, and $p_{Y|Z}(y|z) = p_{Y|X}(y|G(z))$, that is $G$ is considered as a deterministic transformation from the latent to the image space. 
The original approach introduced in \cite{Bora2017} is efficient for solving severely ill-posed inverse problems on images lying on a low-dimensional manifold, such as centered close-ups on human faces. Unfortunately, 
the manifold constraint $x = G(z)$ too strongly restricts the solution space when dealing with images lying on a high-dimensional manifold, which is the case for highly diverse and/or non-structured images,
or when solving less severely ill-posed inverse problems. 
To address this limitation, \cite{Dhar2018,Dean2020} permit small deviations from the generative manifold while \cite{Gonzalez2022,Duff2022,Duff2023} propose to jointly optimize $x$ and $z$ using splitting algorithms.
Solving \cref{eq_bora} using Normalizing Flows (NF) \cite{Dinh2015} is also an option as it ensures the accessibility of any element in the image space \cite{Asim2020,Oberlin2021}.
Even so, the quality of the solution remains highly dependent on the generative model.
Unfortunately, the use of very deep state-of-the-art generative models makes the optimization of \cref{eq_bora} problematic. Indeed, the gradient descent often gets trapped in poor local minima \cite{Daras2020} and it can be regarded as too computationally demanding for restoring a single image. Moreover, the training of very deep generative models require\toadd{s} a lot of ideal images, which might not be accessible in some application contexts. 
\toadd{To} circumvent gradient descent, \cite{Gonzalez2022} uses the encoder of a variational autoencoder (VAE) as a stochastic approximate of the generative model posterior 
in an alternate optimization scheme on $(x,z)$, while \cite{Prost2023} obtains promising performance by extending this approach to hierarchical VAEs \cite{Soenderby2016}, at the cost of additional image restoration hyperparameters. \toadd{Alternatively, PULSE \cite{Menon2020} uses  
gradient descent over 
a high probability hyper-sphere in the latent space. This allows to reduce the blur induced by the 
quadratic latent regularization associated with the 
Gaussian latent prior.}

All the above mentioned methods only provide a single point estimate of the inverse problem solution. However, many applications require an estimation of the confidence behind this solution.
To this end, several Bayesian methods enable to sample the posterior distribution $p_{X|Y}(x|y)$ of the inverse problem solution.
Some methods yield stochastic solutions to the inverse problem, relying, for instance, on the implicit prior provided by Gaussian denoisers \cite{Kadkhodaie2021} or exploiting diffusion model properties \cite{Zhu2023}\toadd{\cite{Kawar2022}}.
However, posterior sampling is then computationally expensive 
since the inverse problem must be solved for each sample generation. In this regard, Markov Chain Monte Carlo (MCMC) can lead to more effective techniques. 
In particular, \cite{Durmus2018} proposes to use Unadjusted Langevin Algorithm (ULA) to solve imaging inverse problems.  Building on this, PnP-ULA \cite{Laumont2022a} approximates the log-likelihood gradient in ULA by employing a Gaussian denoiser in a PnP 
framework. NF-ULA\cite{Cai2023} directly approximates this gradient using normalizing flows while \cite{Holden2022} uses a MCMC sampling scheme in the latent space of a generative model. 
Many of these methods offer theoretical convergence guarantees towards the actual posterior distribution. However\, it requires in practice many iterations which makes the restoration computationally heavy, despite
recent attempts to speed up Markov chain convergence \cite{Coeurdoux2023,Pereyra2020}.
\toadd{Interestingly, variational inference (VI) 
\cite{Likas2004} consists in 
explicitly approximating the posterior distribution by 
a parametric distribution. This allows for 
a fast sampling procedure. However, the existing VI deep-learning-based methods essentially learn, in a supervised manner, a neural network that fits 
the variational parameters to solve a given 
inverse problem \cite{Yue2024,Blei2017,Tonolini2020}}. 

Focusing on latent optimization methods, this paper addresses the 
issues arising when dealing with diverse or non-structured images. Moreover, taking benefit of the latent framework, the proposed approach offers efficient means 
for posterior sampling.
Hence, our contribution is twofold. First, we propose compressive autoencoders (CAEs) \cite{Balle2017} as an alternative to state-of-the-art generative models for latent optimization methods.
We consider in particular CAEs with a hyperprior \cite{Balle2018}, which yield excellent results in compression and can be seen as VAEs with an adaptive latent prior. In addition, they 
are significantly smaller than state-of-the-art generative models.
On one hand, their adaptability allows \toadd{them} to properly regularize inverse problems on non-structured images. On the other hand, their light structure makes the latent optimization using gradient descent \cite{Bora2017} both scalable and effective.
Second, we introduce the Variational Bayes Latent Estimation (VBLE) algorithm which estimates the latent posterior $p_{Z|Y}(z|y)$ by variational inference, leveraging CAE characteristics to design a simple yet efficient approximation of $p_{Z|Y}(z|y)$. 
In this way, VBLE enables to estimate the posterior distribution 
with negligible additional computational cost.

We conduct a comprehensive set of experiments on FFHQ \cite{Karras2019} and BSD \cite{Martin2001} datasets, comprising three different inverse problems:  
deblurring, single image super\toadd{-}resolution (SISR), and inpainting. Additionally, we perform a thorough study to demonstrate the relevance of VBLE posterior distribution. 
The proposed approach yields competitive image restoration results compared to state-of-the-art methods. Furthermore, it outperforms other posterior sampling methods in terms of computation time and GPU load, while yielding consistent abilities for posterior sampling and uncertainty quantification compared to state-of-the-art baselines.


\section{Background}

\subsection{Variational autoencoders}
\label{sec:vae}

\noindent VAEs are composed of an encoder, known as the inference model as it approximates the unknown posterior distribution, and of a decoder used as a generative model \cite{DiederikPKingma2014}. Both are neural networks, whose weights, $\phi$ and $\theta$ respectively, are learned using variational inference \cite{Blei2017}.
Let the image $x$ depend on a latent variable $z$ according to the following generative model: 
\begin{equation}
    p_\theta (x,z) = p_\theta(x|z) p_\theta (z). \label{eq:genmodel}
\end{equation}
$p_\theta(z)$ is typically a simple distribution, often chosen as $\mathcal{N}(0,I)$, while $p_\theta(x|z)$ represents the distribution learned by the decoder.
The posterior distribution, defined as
\begin{equation}
    p_\theta(z|x) = \frac{p_\theta(x|z) p_\theta (z)}{\int_{z} p_\theta(x|z) p_\theta (z)}
\end{equation}
is often intractable and is approximated by $q_\phi(z|x)$, the inference model produced by the VAE encoder.

The weights $\theta$ and $\phi$ are learned during the training of the VAE by maximizing the Evidence Lower BOund (ELBO):

\begin{equation}
    \mathcal{L}_{\theta,\phi} (x) = \log p_\theta(x) - KL(q_\phi(z|x) || p_\theta(z|x))
\end{equation}
where KL denotes the Kullblack-Leibler divergence, measuring the distance between distributions. Hence, by maximizing the ELBO on a dataset $\mathcal{D}$ containing several images $x$, the log-likelihood of $\mathcal{D}$ is maximized, while the distance between the approximate and the true posterior distributions is minimized. However, this version of the ELBO is intractable 
and can be rewritten as follows:

\begin{equation}
    \mathcal{L}_{\theta,\phi} (x) = \mathbb{E}_{q_\phi (z|x)} \big[ \log p_\theta(x|z) \big] - KL\big(q_\phi(z|x) || p_\theta(z)\big). \label{elbo}
\end{equation}
It consists of two terms. The first one is a data fidelity term which ensures a good reconstruction of the input $x$ by the autoencoder. The second one encourages the latent space to match the shape of the prior distribution $p_\theta(z)$. Leveraging the reparameterization trick \cite{DiederikPKingma2014}, Stochastic Gradient Variational Bayes (SGVB) estimates can be derived from the ELBO to update $(\theta,\phi)$.

\subsection{Variational compressive autoencoders}
\label{sec:caes}

\begin{figure}[t]
    \centering
    \includegraphics[width = 0.5\textwidth]{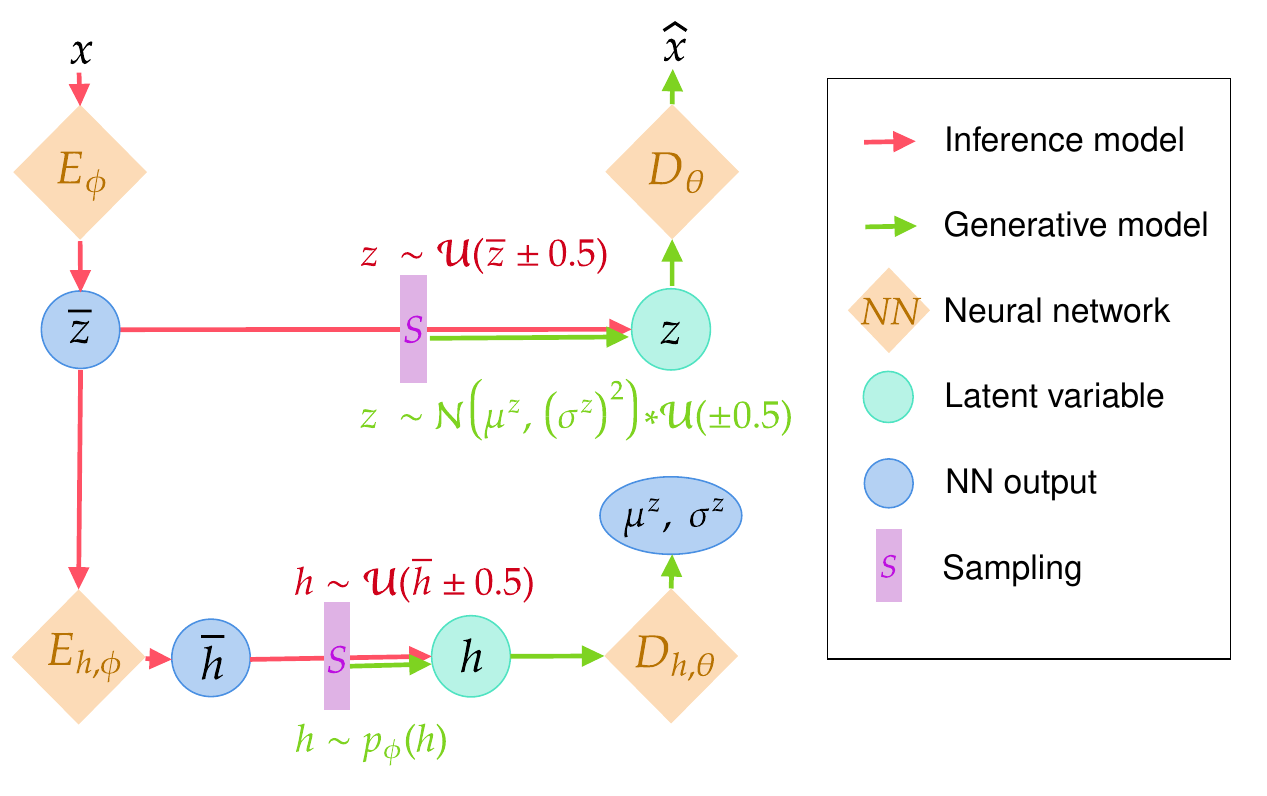}
    \caption{\footnotesize{Structure of a compressive autoencoder with hyperprior as in \cite{Balle2018}. All mentionned distributions are factorized, in particular $\mathcal{N}(\mu^z,(\sigma^z)^2)*\mathcal{U}(\pm 0.5) = \prod_k \big[ \mathcal{N}(\mu^z_k,(\sigma^z_k)^2)*\mathcal{U}(-0.5,0.5) \big]$.}}
    \label{aecomp}
\end{figure}
\noindent Machine learning and especially deep learning have raised a significant interest in the lossy image compression community. In 
the widespread transform coding framework, the image is transformed, the obtained image representation is quantized and subsequently compressed using a lossless entropy encoder
\cite{Wintz1972}.
The more decorrelated and sparse the transform is, the more effective the compression becomes.
Several reference model-based methods use the wavelet transform \cite{Skodras2001} while state-of-the-art deep learning methods use data-dependent transforms provided by the encoder part of a compressive autoencoder (CAE) \cite{Balle2017,Balle2018,Minnen2018,Cheng2020}.
CAEs are learned by minimizing a rate-distortion trade-off $\mathcal{L} = \mbox{ Rate}+\alpha \times \mbox{Distortion }$ \cite{Balle2017,Balle2018}.
On one hand, the quantization introduces distortion between the original and the decompressed image. This distortion is typically measured by the Mean Squared Error (MSE) between the network input and output. On the other hand, the entropy encoder design requires a prior distribution or so-called entropy model $p$ for the quantized latent representation, denoted by $z$ in the following. However, $z$ actual distribution, denoted by $q$, generally differs from $p$. The rate is approximated by the Shannon cross\toadd{-}entropy: 
$\mbox{Rate }=\mathbb{E}_{z \sim q} [-\log_2 p(z)]$.  This value is minimized when $p=q$, that is when the rate corresponds to the information entropy.

A specificity of CAE training comes from the fact that quantization is non-differentiable. It is thus approximated during training by the addition of a uniform noise, which classically models quantization noise \cite{Bovik2010}.
This noise is akin to the noise introduced during the training of a VAE in the generative framework. In the following, we denote by  $\bar z$ the representation before the quantization. 
The latent variable $z$ is thus a noisy version of $\bar z$.
\begin{figure*}[t]
    \centering
    \includegraphics[width = 1\textwidth]{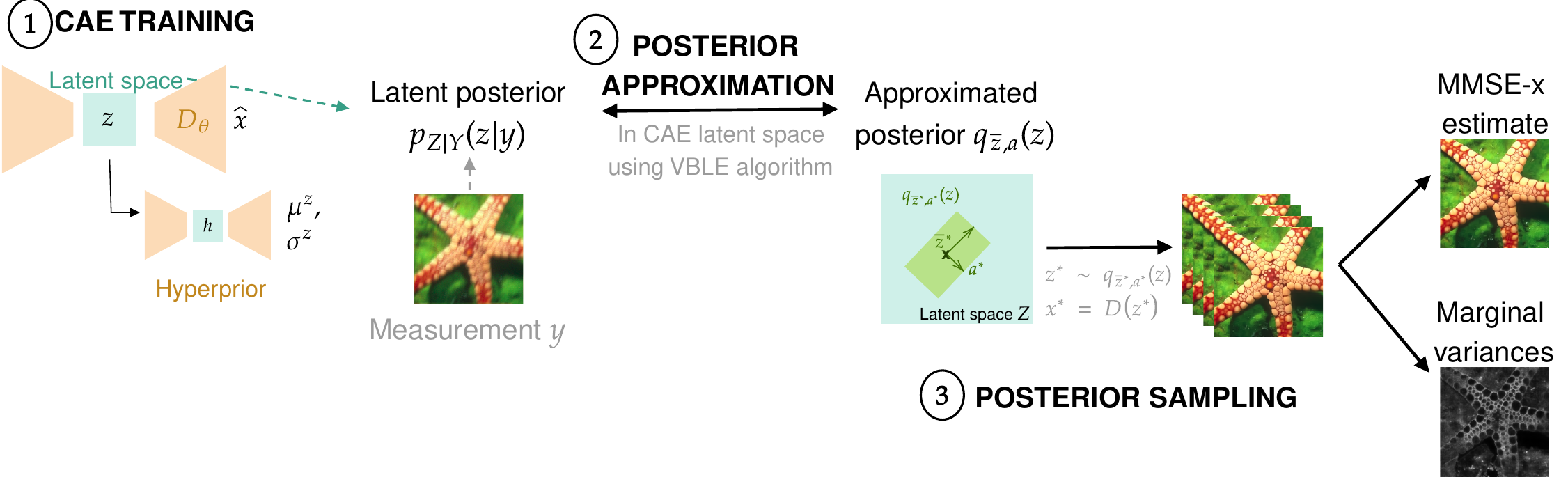}
    \caption{\footnotesize{Overview of the image restoration process. First, a compressive autoencoder (CAE) is trained on a dataset of ideal images. Then, to restore a degraded image $y$, the latent posterior $p_{Z|Y}(z|y)$ is approximated using variational inference, and finally,  the approximated posterior $q_{\bar z, a}(z)$ is sampled in order to compute an MMSE estimate of the solution as well as uncertainties.}}
    \label{fig_abstract}
\end{figure*}
Furthermore, state-of-the-art CAEs incorporate a hyperprior \cite{Balle2018,Minnen2018,Chen2021}. This additional autoencoder takes $\bar z$ as input and estimates $z$'s mean $\mu^z$ and standard deviation $\sigma^z$, 
as illustrated in \cref{aecomp}. The prior on $z$ is then defined as the $\bar z$-dependent factorized distribution $\prod_k \mathcal{N}(\mu^z_k,(\sigma^z_k)^2)$  
convolved by a uniform distribution to align with quantization. 
The hyperprior introduces another latent variable $h$, 
of \toadd{a} much smaller dimension than $z$. $h$ is also 
quantized, entropy coded, and transmitted. 

CAEs can be formulated as VAEs 
\cite{Balle2017, Balle2018}, with particular generative and inference models. 
Consider the following generative model, corresponding to a CAE with a hyperprior:
\begin{align}
    p_\theta(x,z,h) &= p_\theta(x|z)p_\theta(z|h)p_\theta(h) \\
    \mbox{with } p_\theta(x|z) &= \prod_k \mathcal{N}\big(x_k;D_\theta(z)_k,\frac{1}{2\alpha\log(2)}\big) \nonumber\\
    p_\theta(z|h) &= \prod_k \Big[ \mathcal{N}(\mu^z_k,(\sigma^z_k)^2)*U\big(z_k;\, [-\frac{1}{2},\frac{1}{2}]\big)\Big] \nonumber \\
    p_\theta(h) &= \prod_k p_{\psi}(h_k) \nonumber
\end{align}
where $D_\theta$ represents the decoder part of the autoencoder, and $p_{\psi}$ denotes a factorized parametric prior with weights $\psi$ learned during training. Consider also the following inference model:
\begin{align}
    q_\phi(z,h|x) &= q_\phi(z|x, h)q_\phi(h|x) \\
    \mbox{with } q_\phi(z|x, h) &= \prod_k \mathcal{U}(z_k;[\bar z_k-\frac{1}{2},\bar z_k+\frac{1}{2}]),  \nonumber\\
    q_\phi(h|x) &= \prod_k \mathcal{U}(h_k;[\bar h_k-\frac{1}{2},\bar h_k+\frac{1}{2}]). \nonumber
\end{align}
Then, the ELBO from \cref{elbo}, expressed for a hierarchical VAE with two latent variables, corresponds to the rate-distortion 
loss, up to a $\log(2)$ factor:
\begin{align}
    \mathcal{L}(x) &= \mathbb{E}_{q_\phi(z,h|x)}\big[\log q_\phi(z,h|x) - \log p_\theta(x|z,h) \nonumber \\ 
    & \quad \quad \quad \quad \quad \quad - \log p_\theta(z,h) \big] \\
    &\propto 0 + \log(2) (\alpha \mbox{Distortion}(x, z) + \mbox{ Rate}(z,h)). \nonumber
\end{align}
Note that $\alpha$ controls the rate-distortion tradeoff, which is similar to the KL-data fidelity tradeoff for VAE that is generally controlled by a $\gamma$ parameter assuming a Gaussian decoder of variance $\gamma^2$ \cite{BinDai2019}.

Furthermore, 
state-of-the-art CAEs combine a second autoencoder with an autoregressive part to design an improved hyperprior \cite{Minnen2018, Cheng2020}. In this case, CAEs cannot be entirely seen as VAEs, although their formulation stay\toadd{s} close.

Finally, CAEs are powerful neural networks, while remaining often scalable, as they are to be used in embedded systems \cite{Oliveira2022}. 
In particular, these networks use a parametric activation function, known as Generalized Divisive Normalization \cite{Balle2015} (GDN), which, compared to classical activation functions, affords an equivalent approximation capacity for natural images with far shallower networks.

\section{Proposed method}

\subsection{Regularization with compressive autoencoders}
\label{sec:caereg}

\noindent We claim that CAEs, introduced in \cref{sec:caes}, are good candidates to be used when restoring images with latent optimization methods.
First, they can be viewed as VAEs and may be employed similarly. Second, they remain relatively light as their structure has been optimized for embedded image processing, enabling the use of gradient descent to compute the MAP-z estimate of \cref{eq_bora}. Last, the hyperprior offers a flexible $z$-adaptive latent prior that better models the true latent distribution than typical VAE priors. 
Therefore, under the formalism of \cref{sec:caes} summarized in \cref{aecomp}, the MAP-z estimate for CAEs can be formulated as: 
\begin{align}
    z^* = \arg\min_{z} 
    \toadd{- \log p_{Y|Z}(y|z)} + \lambda \mbox{ Rate}\toadd{_\theta}(z,E_{h,\phi} (z)) \label{eq_ir1}
\end{align}
where $E_{h,\phi} (z)$ denotes 
the output of the hyperencoder, $\mbox{Rate}\toadd{_\theta}(z,E_{h,\phi} (z))=- \log_2 p_\theta(z,E_{h,\phi} (z))$
\toadd{ and, for the inverse problem $y = Ax + w$ defined in the introduction, }
$- \log p_{Y|Z}(y|z) \propto ||A D_\theta(z) - y||_2^2$.
Note that, in a proper MAP framework, the optimization should be with respect to both $z$ and $h$, 
and the second term in \eqref{eq_ir1} should rather be $ \lambda  \mbox{ Rate}_\theta(z,h)$. As $h$ is low-dimensional and does not significantly impact the rate, we choose the approximation $h = E_{h,\phi} (z)$. We show, in the ablation study provided in \cref{sec:ablation}, that this approximation does not impact the algorithm performance.

\subsection{Variational Bayes Latent Estimation (VBLE)}
\label{sec:vble}

\begin{figure}[t]
    \centering
    \includegraphics[width = 0.5\textwidth]{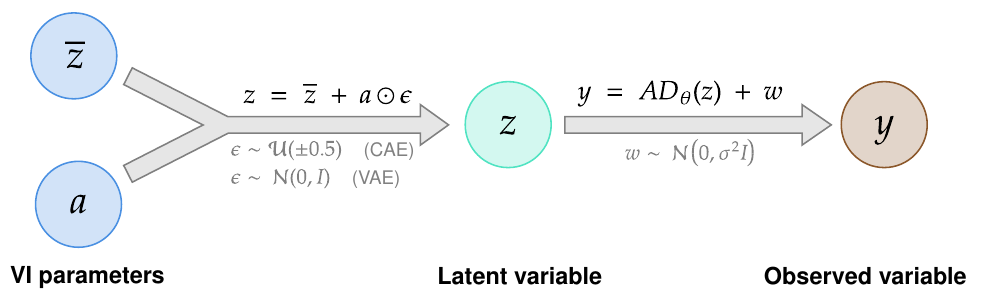}
    \caption{Variational Bayes Latent Estimation (VBLE) graph. VI stands for Variational Inference.}
    \label{vi_graph}
\end{figure}

\noindent We still assume that $p_{Y|Z}(y|z) = p_{Y|X}(y|D_\theta(z))$. Recall that Bora et al. \cite{Bora2017} method to solve inverse problems yields a MAP estimate $\arg \max_{z} p_{Z|Y}(z|y)$ in the latent space. Unlike this deterministic approach resulting in a point estimate, we wish to approximate $p_{Z|Y}(z|y)$ through variational inference \toadd{(VI)}, leveraging VAEs and CAEs latent structure to propose a simple posterior parameterization. Hence, we present a stochastic version of \cref{eq_ir1}, termed Variational Bayes Latent Estimation (VBLE), which is illustrated in \cref{vi_graph}. It consists of performing 
\toadd{VI} with the following parametric families:
\begin{align}
    \toadd{\mathcal{E}}_{\bar z,a} &= \left\{ q_{\bar z,a}(z) \big| \bar z,a \in \mathbb{R}^{C\times M\times N}, a>0 \right\} \label{eq:paramfam} \\
    \text{ with } q_{\bar z,a}(z) &= \prod_k \mathcal{U}(z_k;[\bar z_k-\frac{a_k}{2},\bar z_k+\frac{a_k}{2}]) \quad \text{(CAE case),} \nonumber \\
    q_{\bar z,a}(z) &= \prod_k \mathcal{N}(z_k;\bar z_k, a_k^2) \quad \text{(VAE case).} \nonumber
\end{align}
Note that we detail both CAE and VAE frameworks, as VBLE can be applied with both of them.  
These parametric families are based on the same distribution 
\toadd{as} the inference encoder distribution $q_\phi(z|x)$, \textit{i.e.} uniform for CAEs and Gaussian for VAEs. Their parameters are the mean latent representation $\bar z$, and an additional $a$ parameter with the same dimensions as $\bar z$, which models the uncertainty related to each coefficient of $z$.
A representation of the posterior $q_{\bar z,a}(z)$ is given in \cref{fig_abstract}. The choice of a posterior belonging to the same distribution family as the encoder posterior $q_\phi(z|x)$ is a simple yet relevant choice. Indeed the encoder is trained so that sampling $q_\phi(z|x)$ provides the most likely latent variables given an input image $x$. Here, we aim to find the most likely latent variables given a measure $y$. Hence, adopting a unimodal distribution similar to the encoder posterior, however with \toadd{an }optimized variance, is a natural choice. \toadd{Experiments in \cref{sec:uncertainty} demonstrate that this choice is beneficial to VBLE posterior sampling ability.} 
Then, employing a similar framework as the VAE training procedure described in \cref{sec:vae}, we maximize the ELBO:
\begin{align}
    \arg \max_{\bar z,a} \mathcal{L}_{\bar z,a} &= \arg \max_{z,a} \mathbb{E}_{q_{\bar z,a} (z)} \big[ \log p_{Y|Z}(y|z) \nonumber \\
    &+ \log p_\theta (z) - \log q_{\bar z,a}(z) \big]  \label{eq_ir2}
\end{align}
with $ \log p_\theta(z)$ being the log-likelihood of the latent prior, that is, in particular, $\log \mathcal{N}(z;0,I)$ for a traditional VAE \toadd{as introduced in \cref{eq:genmodel}}, and $\log(2) \mbox{ Rate}\toadd{_\theta}(z,E_{h,\phi}(\bar z))$ for a CAE with a hyperprior, as introduced in \cref{eq_ir1}. Note that $\log q_{\bar z,a}(z) = - \sum_k \log a_k$ up to a constant in both Gaussian and uniform cases. 
We can then use the following reparameterization, for $z \sim q_{\bar z,a}(z)$,
\begin{equation}
    z = \left\{ 
    \begin{array}{ll}
        \bar z + a \odot \epsilon \mbox{ with } \epsilon \sim \prod_k \mathcal{U}(\epsilon_k;\, [-\frac{1}{2}, \frac{1}{2}]) & \mbox{ (CAE case)}\\
        \bar z + a \odot\epsilon \mbox{ with } \epsilon \sim \mathcal{N}(0,I) & \mbox{ (VAE case)}
    \end{array}
    \right. \nonumber
\end{equation}
where $\odot$ denotes the Hadamard, point-wise product between two tensors. 
Hence, the expectation in \cref{eq_ir2} can be taken on $\epsilon$ using the reparametrization trick \cite{DiederikPKingma2014}. Therefore, a\toadd{n} SGVB estimate of the ELBO can be derived, leading to \cref{alg_ir}. 

\begin{algorithm}
\small{
\caption{Variational Bayes Latent Estimation}\label{alg_ir}
\begin{algorithmic}
    \Require \toadd{$y$ a degraded image,} $\bar z_0\in \mathbb{R}^{C\times M\times N}$, $a_0 \in \mathbb{R}^{C\times M\times N}=(1)_{i,j,l}$, $k=0$, $\eta > 0$
    \While{not \textit{convergence}}
        \State $z \sim q_{\bar z_k,a_k}(z)$
        \State $\begin{pmatrix} \bar z_{k+1}\\ a_{k+1} \end{pmatrix} = \begin{pmatrix} \bar z_{k}\\ a_{k} \end{pmatrix} - \eta \nabla_{\bar z,a}\Big[ - \log p_{Y|Z}(y|z) - \log p_\theta (z) + \log q_{\bar z,a}(z) \big) \Big]$
        \State $k = k + 1$
    \EndWhile \\
    \Return $(\bar z^*, a^*) = (\bar z_k,a_k)$
\end{algorithmic}
}
\end{algorithm}

Hence, given the optimal parameters $(\bar z^*,a^*)$, the distribution $q_{\bar z^*,a^*}(z)$ is supposed to approximate the posterior $p_{Z|Y}(z|y)$. Then, two point estimates can be derived for the restored image:
\begin{align}
x^*_{MMSE-z} &= D_\theta(\bar{z}^*), \label{eq:mmsez}\\
x^*_{MMSE-x} &= \frac{1}{L} \sum_{i=1}^L D_\theta(z_i) \label{eq:mmsex}
\mbox{ with } z_i \sim q_{\bar z^*,a^*}(z_i). 
\end{align}
Both MMSE-x and MMSE-z estimates are relevant and can be used in practice. \toadd{Note that this variational framework does not increase the restoration time compared to its deterministic counterpart MAP-z \cite{Bora2017}. 
Indeed, after convergence of VBLE, sampling from the approximate posterior is fast since it only requires forward passes through the decoder, which can be parallelized within batches.}

\subsection{\toaddminor{Discussion and} position to related works}


\toadd{Regarding latent optimization methods, 
most of them consider
point estimates 
\cite{Bora2017, Gonzalez2022, Duff2022}. 
Yet, 
several approaches have proposed posterior sampling or ensemble methods that enable uncertainty quantification. 
 It may be through 
sampling the latent posterior using MCMC \cite{Holden2022,Tezcan2022} or, as in the case of PULSE \cite{Menon2020}, by running a spherical gradient descent multiple times with different random initializations. 
\toaddminor{PULSE does not provide samples of the latent posterior, but several MAP estimates. On the contrary, \cite{Holden2022,Tezcan2022} sample from the latent posterior. In particular, in \cite{Holden2022}, the latent posterior is defined exactly as in VBLE, that is $p_{Z|Y}(z|y) \propto p(z) p_{Y|X}(y|D_\theta(z))$, thus the two 
methods are expected to produce similar posterior samples.}
Compared to \toaddminor{those methods}, VBLE is a Variational Inference (VI) method, \toaddminor{that only approximates the posterior distribution, but with a single latent optimization of the variational parameters.} This key difference leads to a significantly reduced computational load for VBLE compared to \toaddminor{MCMC methods \cite{Holden2022,Tezcan2022} and PULSE \cite{Menon2020}}, as drawing several samples requires several latent optimizations with PULSE, and typically more than $10^6$ iterations with MCMC algorithms.} \toaddminor{Furthermore, to transform the latent samples into images, we consider a deterministic decoder, that is $p_\theta(x|z)=\delta_{D_\theta(z)}(x)$ while \cite{Tezcan2022} obtains image samples by sampling $p(x|z,y) \propto p(x|z)p(y|x)$. By doing so in VBLE, \cref{alg_ir} would remain unchanged, but the decoding, in particular of Equations \ref{eq:mmsez},\ref{eq:mmsex} would be modified: the images would be potentially a bit less blurry. However, our main concern is that the decoder distribution $p_\theta(x|z) = \mathcal{N}(x;D_\theta(z),\gamma^2I)$ classically used in VAEs and CAEs is not well adapted as the fixed variance is not informative enough. Alternative approaches to improve the decoder variance \cite{Duff2023} could be used. As this is not straightforward to apply, this interesting perspective is left for future work.
}

Concerning 
\toadd{VI} methods, they have been widely used to approximate posterior distributions \toadd{of inverse problem solutions} \cite{Blei2017, Tonolini2020}. \toadd{However, }
the approaches which combine 
\toadd{VI} with deep generative modeling are essentially designed in a (semi-)supervised manner to solve a specific inverse problem \cite{Tonolini2020,Gao2022,Goh2019}, while VBLE is able to solve different inverse problems with the same network. In \toadd{\cite{Cheng2023}, denoising is performed through score prior-guided deep VI, VI being used 
to approximate the noise model parameters. This requires optimizing the weights of a convolutional neural network (CNN) that predicts the noise parameters for each input image $y$ to be restored. In contrast, since VBLE assumes that the noise and the degradation are known, it only uses VI to approximate the latent posterior. For that purpose, the variational parameters are directly optimized using gradient descent. Hence, up to our knowledge, VBLE is the first latent optimization method which explicitly estimates the latent posterior with variational inference to solve general inverse problems. 
Moreover, the uniform posterior shape for VBLE with CAE, which mimics the encoder posterior, is novel and yields remarkable results for solving imaging inverse problems, as will be shown in the next section.}


Regarding the 
use of CAEs as regularizers, 
the relationship between CAEs and VAEs has been established in the literature \cite{Balle2017,Mentzer2020}. 
But, 
\toadd{as far as we know}, 
this is the first time that CAEs are used as priors for image restoration tasks. 
Although CAEs are not as deep as state-of-the-art hierarchical VAEs and may not exhibit the same performance for image generation, we believe that they make appropriate priors for image restoration. In particular, the parameter $\alpha$, which tunes the rate-distortion trade-off, provides additional flexibility and can be adjusted for a given inverse problem.


\section{Numerical experiments}

\noindent In this section, we evaluate the performance of VBLE using CAEs. 
With these experiments, our purpose is twofold. First, we aim to show that VBLE, in particular with the use of CAEs, yields state-of-the-art results for image restoration tasks. Second, we wish to demonstrate the huge interest of the proposed method for posterior sampling, which can be crucial for many applications.

In that purpose, we first detail the experimental setup, and then present image restoration results on two datasets. The first one, FFHQ, is highly structured and thus well-suited for generative models, whereas the second one, BSD, contains natural images with more diversity. 
Then, we assess the quality of our proposed approximate posterior both quantitatively and qualitatively compared to state-of-the-art posterior sampling baselines.
Finally, we conduct an ablation study about our method and its variants. 

\subsection{Experimental setup}

\paragraph{Inverse problems} All experiments in this section are performed on subsets of BSD500 \cite{Martin2001} and FFHQ \cite{Karras2019} test datasets, both subsets composed of 47 images of size $256 \times 256$. VBLE and the baselines are evaluated on three inverse problems: deblurring, single image super\toadd{-}resolution (SISR), and inpainting. For deblurring, two Gaussian kernels of standard deviation $\sigma_{blur} \in  \{1,3\}$ are tested as well as a motion blur kernel, for several Gaussian noise levels.
We consider the noiseless case for SISR and inpainting with random masks, and a small amount of noise is added for inpainting with structured masks. 
SISR $\times 2$ and $\times 4$ is conducted with bicubic downsampling. We employ a $50 \%$ random mask for inpainting experiments with random masks. For inpainting with structured masks, custom masks are designed on 5 test images of each dataset.


\paragraph{Metrics} We use 
\toadd{four} metrics to measure the discrepancy between the restored and ground truth images: Peak Signal-to-Noise Ratio (PSNR), Structural SIMilarity \cite{Wang2004} (SSIM), Learned Image Patch Similarity~\cite{Zhang2018} (LPIPS) \toadd{and Fréchet Inception Distance (FID) \cite{Heusel2017}}. The PSNR derives from the pixel-wise mean-squared error, while SSIM and LPIPS are, respectively, classical and deep learning-based perceptual metrics. \toadd{The FID is a no-reference perceptual metric. 
For each inverse problem, we evaluate the FID after extracting 49 patches of size $64\times64$ from each restored image.}

\paragraph{Compressive autoencoder architectures} We employ pretrained networks from the compressAI \cite{Begaint2020} library at different bitrates. 
We choose two models, from Minnen et al. \cite{Minnen2018}, denoted as mbt, and Cheng et al.\cite{Cheng2020}, denoted as cheng. mbt is a CAE with a hyperprior combining a second autoencoder with an autoregressive model, but its encoder and decoder structures remain simple CNNs.
cheng retains the same structure as mbt, but with more elaborate encoders and decoders as they possess self-attention layers, yielding state-of-the-art compression results. 
For mbt and cheng models, two structures, a light and a bigger one, exist \cite{Begaint2020}, which are used respectively for low and high bitrates. For mbt, \toadd{a} high bitrate structure is used for $\alpha \geq 0.013$, with $\alpha$ the bitrate parameter, and \toadd{a} low bitrate structure for $\alpha \leq 0.0067$. For cheng, we have chosen the high bitrate structure for high and low bitrates, as it provides the best image restoration results. The number of parameters for each network is given in \cref{tab:nparams}.
The networks from compressAI are pretrained on natural images. We finetune them for about 100k iterations on BSD and FFHQ train datasets.

\begin{table*}[ht]
    \centering
    \footnotesize
\begin{tabular}{llccccccccccc}
    \hline
    \textbf{Deblur - $\sigma=7.65$} & \multicolumn{4}{c}{\textbf{Gaussian}, $\sigma_k=1$} & \multicolumn{4}{c}{\textbf{Gaussian}, $\sigma_k=3$} & \multicolumn{4}{c}{\textbf{Motion}} \\ \cline{2-13} 
    \textbf{Method} & PSNR $\uparrow$ & SSIM $\uparrow$ & LPIPS $\downarrow$ & FID $\downarrow$ & PSNR $\uparrow$ & SSIM $\uparrow$ & LPIPS $\downarrow$ & FID $\downarrow$ & PSNR $\uparrow$ & SSIM $\uparrow$ & LPIPS $\downarrow$ & FID $\downarrow$ \\ \hline
    VBLE wCAE & \toadd{\textbf{34.48}} & \toadd{\textbf{0.9250}} & \toadd{\underline{0.1473}} & \toadd{38.03} & \toadd{\textbf{29.57}} & \toadd{\textbf{0.8214}} & \toadd{\underline{0.2859}} & \toadd{74.12} & \toadd{\textbf{32.22}} & \toadd{\textbf{0.8800}} & \toadd{\underline{0.1933}} & \textbf{\toadd{40.70}} \\
    MAP-z wCAE & 34.17 & 0.9152 & 0.1684 & \toadd{48.88} & \underline{29.44} & \underline{0.8176} & 0.2977 & \toadd{92.65} & \underline{32.01} & 0.8717 & 0.2441 & \toadd{64.08} \\
    VBLE wCAE (mbt) & \underline{34.21} & \underline{0.9177} & 0.1600 & \toadd{43.44} & 28.70 & 0.7686 & 0.3212 & \toadd{102.96} & 31.97 & \underline{0.8741} & 0.2149 & \toadd{48.15} \\ \hline
    MAP-z wVAE V1 & 32.78 & 0.8868 & 0.1990 & \toadd{71.43} & 28.72 & 0.7961 & 0.3242 & \toadd{120.15} & 27.95 & 0.6582 & 0.4331 & \toadd{192.81} \\
    MAP-z wVAE V2 & 30.91 & 0.8486 & 0.2688 & \toadd{66.01} & 28.47 & 0.7849 & 0.3263 & \toadd{\underline{79.12}} & 29.44 & 0.8112 & 0.3088 & \toadd{73.55} \\
    \toadd{PULSE} & \toadd{29.99} & \toadd{0.8289} & \toadd{0.2848} & \toadd{64.42} & \toadd{27.83} & \toadd{0.7670} & \toadd{0.3429} & \toadd{81.00} & \toadd{28.36} & \toadd{0.7848} & \toadd{0.3266} & \toadd{74.85} \\
    DiffPIR & 33.92 & 0.9049 & \textbf{0.1368} & \toadd{\underline{36.12}} & 29.41 & 0.8102 & \textbf{0.2691} & \toadd{100.98} & 31.98 & 0.8677 & \textbf{0.1793} & \toadd{\underline{42.50}} \\ 
    \toadd{DDRM} & \toadd{32.22} & \toadd{0.8831} & \toadd{0.1785} & \textbf{\toadd{33.84}} & \toadd{24.74} & \toadd{0.7051} & \toadd{0.2931} & \textbf{\toadd{41.01}} & x & x & x & x \\ \hline
\end{tabular}

\begin{tabular}{llccccccccccc}
    \hline
    \textbf{SISR and Inp.} & \multicolumn{4}{c}{\textbf{SISR} $\times 2$} & \multicolumn{4}{c}{\textbf{SISR} $\times 4$} & \multicolumn{4}{c}{\textbf{Inpainting} (random)} \\ \cline{2-13} 
    \textbf{Method} & PSNR $\uparrow$ & SSIM $\uparrow$ & LPIPS $\downarrow$ & FID $\downarrow$ & PSNR $\uparrow$ & SSIM $\uparrow$ & LPIPS $\downarrow$ & FID $\downarrow$ & PSNR $\uparrow$ & SSIM $\uparrow$ & LPIPS $\downarrow$ & FID $\downarrow$ \\ \hline
    VBLE wCAE & \toadd{\textbf{36.16}} & \toadd{\textbf{0.9468}} & \toadd{0.1251} & \toadd{40.44} & \toadd{\textbf{31.18}} & \toadd{\textbf{0.8670}} & \toadd{0.2290} & \toadd{\underline{56.21}} & \toadd{\underline{35.94}} & \toadd{\underline{0.9584}} & \toadd{0.0638} & \toadd{31.01} \\
    MAP-z wCAE & \underline{36.30} & \textbf{0.9516} & 0.0995 & \toadd{34.59} & \underline{31.08} & \underline{0.8681} & 0.2205 & \toadd{62.91} & 36.91 & 0.9618 & \underline{0.0715} & \toadd{\underline{30.42}} \\
    VBLE wCAE (mbt) & 35.99 & 0.9514 & \underline{0.0956} & \toadd{38.64} & 31.01 & 0.8680 & 0.2258 & \toadd{60.95} & \textbf{37.04} & \textbf{0.9643} & 0.0738 & \toadd{33.73} \\ \hline
    MAP-z wVAE V1 & 34.35 & 0.9311 & \textbf{0.0893} & \toadd{38.51} & 30.24 & 0.8499 & 0.2201 & \toadd{75.74} & 35.21 & 0.9443 & 0.0775 & \toadd{35.00} \\
    MAP-z wVAE V2 & 31.20 & 0.8534 & 0.2592 & \toadd{64.24} & 29.53 & 0.8226 & 0.2765 & \toadd{66.55} & 31.01 & 0.8513 & 0.2660 & \toadd{66.29} \\
    \toadd{PULSE} & \toadd{29.79} & \toadd{0.8203} & \toadd{0.2989} & \toadd{66.85} & \toadd{27.11} & \toadd{0.7395} & \toadd{0.3927} & \toadd{85.81} & \toadd{30.06} & \toadd{0.8300} & \toadd{0.2858} & \toadd{64.37} \\
    DiffPIR & 34.94 & 0.9269 & 0.1089 & \toadd{\underline{34.63}} & 30.71 & 0.8492 & \underline{0.2042} & \toadd{68.34} & 36.08 & 0.9470 & \textbf{0.0653} & \toadd{30.93} \\ 
    \toadd{DDRM} & \toadd{35.81} & \toadd{0.9479} & \toadd{\textbf{0.0640}} & \textbf{\toadd{27.19}} & \toadd{30.80} & \toadd{0.8608} & \toadd{\textbf{0.1749}} & \textbf{\toadd{38.65}} & \toadd{34.43} & \toadd{0.9372} & \toadd{0.0814} & \textbf{\toadd{24.54}} \\ \hline
\end{tabular}
\caption{\footnotesize{FFHQ results on diverse inverse problems. For methods using CAEs, (mbt) means that mbt CAE \cite{Minnen2018} is used, otherwise cheng CAE \cite{Cheng2020} is used. $\sigma$ is the noise level used for deblurring. $\sigma_k$ is the kernel standard deviation for Gaussian deblurring.
    }}
        \label{tab:res_ffhq}
\end{table*}

\paragraph{VBLE parameters} For VBLE using CAEs, the network with the most appropriate bitrate needs to be chosen for each inverse problem. We have trained models at 7 different bitrates. The bitrate parameter $\alpha$ to choose, as well as the regularization parameter $\lambda$, are tuned by a grid search algorithm on a 3\toadd{-}image validation set for each dataset. The other parameters are fixed, in particular, we use Adam optimizer with \toadd{a} learning rate \toadd{of} 0.1 during 1000 iterations for all restoration tasks. Lastly, the MMSE-x estimate\toadd{, computed by averaging 100 VBLE posterior samples,} is used for all comparisons as it yielded the best results. 

\paragraph{Image restoration parameters} To tune image restoration hyperparameters of VBLE and of the baselines, a grid search algorithm has been used on separate 3 image sets of FFHQ and BSD test datasets. The values of all the VBLE parameters for the different problems are given in the appendix, as well as the parameters \toadd{that} have been tuned or fixed for each baseline.

\subsection{FFHQ experiments}
\label{sec:ffhq}

\begin{figure*}[ht]
    \centering
    \addbox{
    \includegraphics[width = 0.9\textwidth]{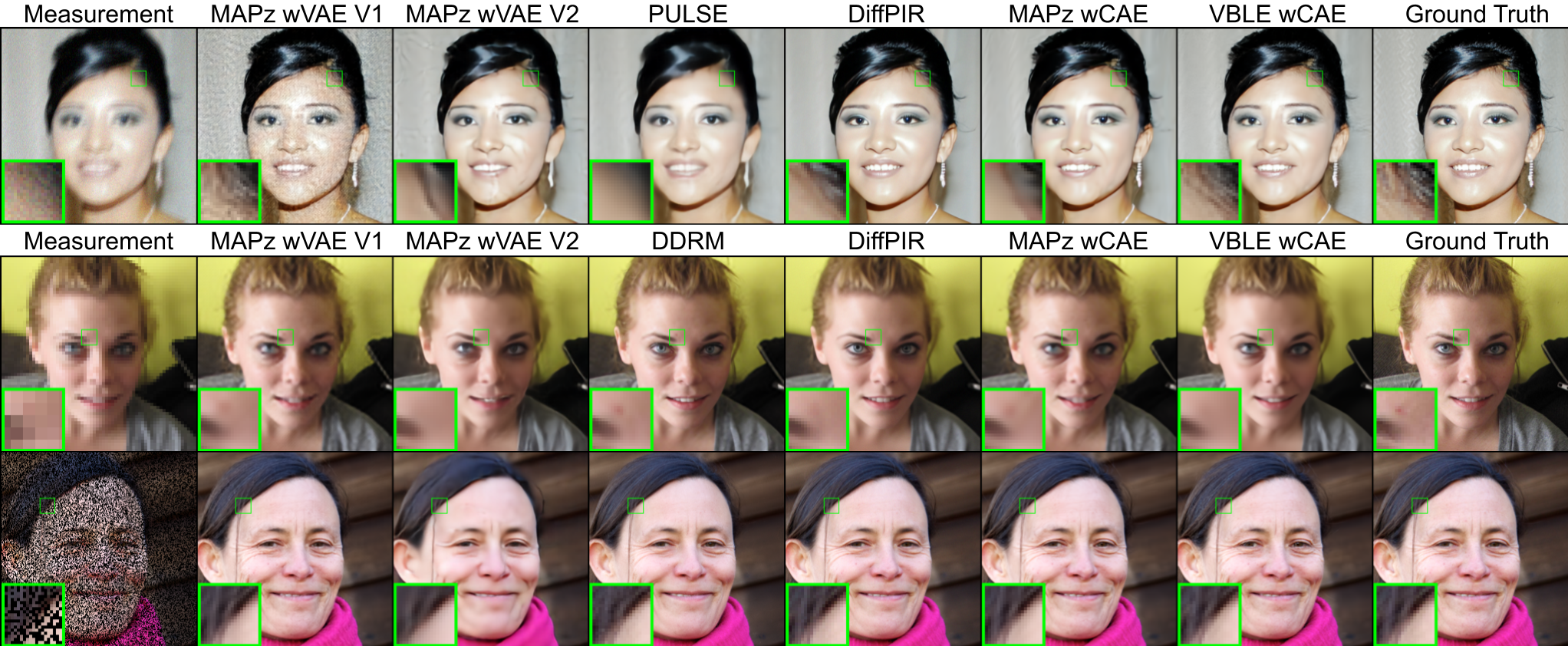}
    }
    \caption{\footnotesize{Visual results on FFHQ. From top to bottom: : motion deblurring ($\sigma=7.65/255$), SISR $\times 4$, inpainting with random masks.}}
    \label{fig:res_ffhq}
\end{figure*}
\begin{table*}[ht]
    \centering
    \footnotesize{
\begin{tabular}{llccccccccccc}
    \hline
    \textbf{BSD} - $\sigma=7.65$ & \multicolumn{4}{c}{\textbf{Deblur} (Gauss., $\sigma_k=1$)} & \multicolumn{4}{c}{\textbf{Deblur} (Gauss., $\sigma_k=3$)} & \multicolumn{4}{c}{\textbf{Deblur} (Motion)} \\ \cline{2-13} 
    \textbf{Method} & PSNR $\uparrow$ & SSIM $\uparrow$ & LPIPS $\downarrow$ & FID $\downarrow$ & PSNR $\uparrow$ & SSIM $\uparrow$ & LPIPS $\downarrow$ & FID $\downarrow$ & PSNR $\uparrow$ & SSIM $\uparrow$ & LPIPS $\downarrow$ & FID $\downarrow$ \\ \hline
    VBLE wCAE & \toadd{\textbf{29.62}} & \toadd{\underline{0.8598}} & \toadd{\underline{0.1901}} & \toadd{61.38} & \toadd{\underline{24.66}} & \toadd{0.6433} & \toadd{\underline{0.3956}} & \toadd{\underline{89.22}} & \toadd{\underline{28.38}} & \toadd{\underline{0.8158}} & \toadd{\underline{0.2356}} & \toadd{\underline{64.91}} \\
    MAPz wCAE & 29.28 & 0.8405 & 0.2215 & \toadd{68.03} & 24.48 & 0.6347 & 0.4193 & \toadd{119.67} & 28.07 & 0.7896 & 0.3111 & \toadd{101.16} \\
    \hline
    PnP-ULA & 27.71 & 0.7809 & 0.2675 & \toadd{67.50} & 23.75 & 0.5913 & 0.4592 & \toadd{114.23} & 27.17 & 0.7402 & 0.2643 & \toadd{71.56} \\
    DPIR & \underline{29.72} & \textbf{0.8623} & 0.2127 & \toadd{73.76} & 24.41 & \textbf{0.6427} & 0.4315 & \toadd{134.92} & \textbf{28.69} & \textbf{0.8285} & 0.2606 & \toadd{83.44} \\
    PnP-ADMM & 28.96 & 0.8456 & 0.2155 & \toadd{67.93} & 24.24 & 0.6333 & 0.4350 & \toadd{131.58} & 28.16 & 0.8045 & 0.3349 & \toadd{127.81} \\
    DiffPIR & 29.26 & 0.8420 & \textbf{0.1832} & \textbf{\toadd{55.34}} & \textbf{24.58} & \underline{0.6405} & \textbf{0.3605} & \toadd{98.93} & 28.15 & 0.7955 & \textbf{0.2230} & \textbf{\toadd{61.19}} \\
    \toadd{DDRM} & \toadd{27.03} & \toadd{0.7763} & \toadd{0.2238} & \toadd{\underline{56.12}} & \toadd{20.32} & \toadd{0.4747} & \toadd{0.4376} & \textbf{\toadd{71.75}} & x & x & x & x \\
    \hline
\end{tabular}

\begin{tabular}{llccccccccccc}
    \hline
    \textbf{BSD} & \multicolumn{4}{c}{\textbf{SISR} $\times$2} & \multicolumn{4}{c}{\textbf{SISR} $\times$4} & \multicolumn{4}{c}{\textbf{Inpainting} (Random)} \\ \cline{2-13} 
    \textbf{Method} & PSNR $\uparrow$ & SSIM $\uparrow$ & LPIPS $\downarrow$ & FID $\downarrow$ & PSNR $\uparrow$ & SSIM $\uparrow$ & LPIPS $\downarrow$ & FID $\downarrow$ & PSNR $\uparrow$ & SSIM $\uparrow$ & LPIPS $\downarrow$ & FID $\downarrow$ \\ \hline
    VBLE wCAE & \toadd{\textbf{29.55}} & \toadd{0.8748} & \toadd{0.1802} & \toadd{65.72} & \toadd{\textbf{25.53}} & \toadd{\textbf{0.7032}} & \toadd{0.3507} & \toadd{99.09} & \toadd{30.21} & \toadd{0.9070} & \toadd{0.1207} & \toadd{56.60} \\
    MAPz wCAE & \textbf{29.65} & 0.8757 & 0.1675 & \toadd{62.53} & \underline{25.38} & \underline{0.6966} & 0.3462 & \toadd{100.87} & 30.45 & 0.9091 & 0.1247 & \toadd{56.25} \\
    \hline
    PnP-ULA & 28.42 & \underline{0.8763} & \underline{0.1659} & \toadd{57.40} & 24.82 & 0.6894 & 0.3730 & \toadd{108.86} & 26.29 & 0.8122 & 0.2497 & \toadd{81.46} \\ 
    DPIR & 29.55 & \textbf{0.8784} & 0.1706 & \toadd{61.35} & 25.18 & 0.6937 & 0.3652 & \toadd{114.41} & \underline{31.58} & \underline{0.9316} & \textbf{0.0659} & \toadd{58.20} \\
    PnP-ADMM & 29.31 & 0.8678 & 0.1907 & \toadd{66.94} & 24.77 & 0.6659 & 0.4227 & \toadd{139.96} & \textbf{31.69} & \textbf{0.9322} & \underline{0.0675} & \toadd{\underline{53.68}} \\
    DiffPIR & 29.20 & 0.8539 & \textbf{0.1645} & \toadd{\underline{57.78}} & 25.03 & 0.6771 & \underline{0.3164} & \toadd{\underline{88.83}} & 30.41 & 0.9018 & 0.0919 & \toadd{55.94} \\
    \toadd{DDRM} & \toadd{28.73} & \toadd{0.8691} & \toadd{\textbf{0.1284}} & \textbf{\toadd{51.27}} & \toadd{25.12} & \toadd{0.6895} & \toadd{\textbf{0.2885}} & \textbf{\toadd{64.57}} & \toadd{29.48} & \toadd{0.8754} & \toadd{0.1268} & \textbf{\toadd{52.59}} \\
    \hline
\end{tabular}
}
 \caption{\footnotesize{BSD results on diverse inverse problems. 
    $\sigma$ is the noise level used for deblurring. $\sigma_k$ is the kernel standard deviation for Gaussian deblurring.}}
    \label{tab:res_bsd}
\end{table*}

\noindent First, we demonstrate the relevance of VBLE using CAEs for restoring highly structured face images from FFHQ dataset. To assess the effectiveness of both the CAE architecture and VBLE algorithm, we provide restoration results with VBLE (denoted as VBLE wCAE in the experiments) et also with its deterministic counterpart MAP-z using the same CAE (denoted as MAP-z wCAE). Results are provided with cheng and mbt CAE structures.
\paragraph{FFHQ Baselines}
We compare our method to \toadd{several} latent optimization approach\toadd{es. First, MAP-z wVAE,} proposed by Bora et al. \cite{Bora2017} \toadd{is the deterministic counterpart of VBLE, which} 
seems a natural baseline. \toadd{We also compare to PULSE \cite{Menon2020}, which performs spherical gradient descent in the latent space. Then, we compare to}
 the state-of-the-art diffusion-based method\toadd{s} DiffPIR \cite{Zhu2023} \toadd{and DDRM \cite{Kawar2022}}. \toadd{We use 100 Neural Function Evaluations (NFEs) for both methods.} DiffPIR's \toadd{and DDRM's} outputs 
\toadd{are} a sample from the posterior distribution $p(x|y)$. Hence, \toadd{they }
can be considered as posterior sampling methods, even if sampling from the posterior requires running the restoration process multiple times. When evaluating image restoration performance, we choose to run DiffPIR \toadd{and DDRM} only once, as is classically done in the literature. \toadd{For DDRM, we do not perform motion blur as it is not implemented in the source code.} For 
\toadd{these baselines}, we use an available FFHQ pre-trained diffusion model \cite{Choi2021}.
For Bora et al. method \cite{Bora2017}, we train two vanilla VAE models on FFHQ, VAE-V1, and VAE-V2. With these models, our aim is to demonstrate the effectiveness of the two latent variable structure specific to CAEs. Hence, these models are vanilla VAEs, that is VAEs with a single latent variable. Their structure has been designed so that it has \toadd{the} same depth, \toadd{same number of }hidden channels and \toadd{same} latent dimension as mbt
with (respectively without) hyperprior in the case of VAE-V1 (respectively VAE-V2). Hence, VAE-V1 encoder possesses 5 convolutional layers and VAE-V2 7 convolutional layers. The latent dimensions of VAE-V1 and V2 correspond to $z$ (first latent variable) and $h$ (second latent variable) dimensions of mbt CAE. Both models have been trained on FFHQ using a learning rate of $1e^{-4}$ during at least 300k iterations. We choose a Gaussian assumption for the decoder, that is $p_\theta(x|z) = \mathcal{N}(D_\theta (z), \gamma^2 I).$ The variance $\gamma^2$ is considered a global parameter and is learned with the other network parameters. \toadd{VAE-V2 model is used for PULSE \cite{Menon2020} baseline.}
Although more elaborate VAEs are likely to perform better as long as their size does not hamper the latent optimization, here we aim to assess the benefit of the CAE architecture, with its specific hyperprior. Hence we choose to compare with VAEs having similar complexity.


\paragraph{Results}
Quantitative results on FFHQ  are provided in \cref{tab:res_ffhq}.
First, VBLE exhibits state-of-the-art results compared to the baselines. Compared to DiffPIR, VBLE exhibits better PSNR and SSIM, \toadd{a similar FID,}  and slightly lower LPIPS scores, which indicates that these methods have a slightly different position in the perception-distortion tradeoff \cite{Blau2018}. \toadd{DDRM outperforms all methods in FID, but is below VBLE and DiffPIR for the other metrics.} Then Bora's method MAP-z with VAEs \toadd{and PULSE} are outperformed by VBLE.
Furthermore, VBLE with CAE yields better metrics than MAP-z with CAE, hence 
the use of VBLE algorithm instead of a deterministic gradient descent not only enables posterior sampling but also improves point estimation results. 
Additionally, visual comparisons are provided in \cref{fig:res_ffhq}. VBLE\toadd{, DDRM } and DiffPIR perform very well on FFHQ. \toadd{Compared to VBLE, the diffusion-based methods yield particularly good reconstruction on details such as eyes and teeth.} \toadd{This can be partly explained by the MMSE-x estimate chosen for VBLE. Indeed, it corresponds to the posterior mean, which tends to be smoother but more faithful to the measurement}. 
\toadd{VAE-V1, which imitates a CAE without hyperprior, seems to poorly restore 
structure while VAE-V2, which has \toadd{the} same depth as a CAE with hyperprior but has a single latent variable, lacks 
high frequencies. This demonstrates the interest of the CAE structure and its hyperprior: its two-latent variable structure achieves retrieving both semantic information and finer details. }

\subsection{BSD experiments}
\label{sec:bsd}

\begin{figure*}[ht]
    \centering
    \addbox{
    \includegraphics[width = 0.9\textwidth]{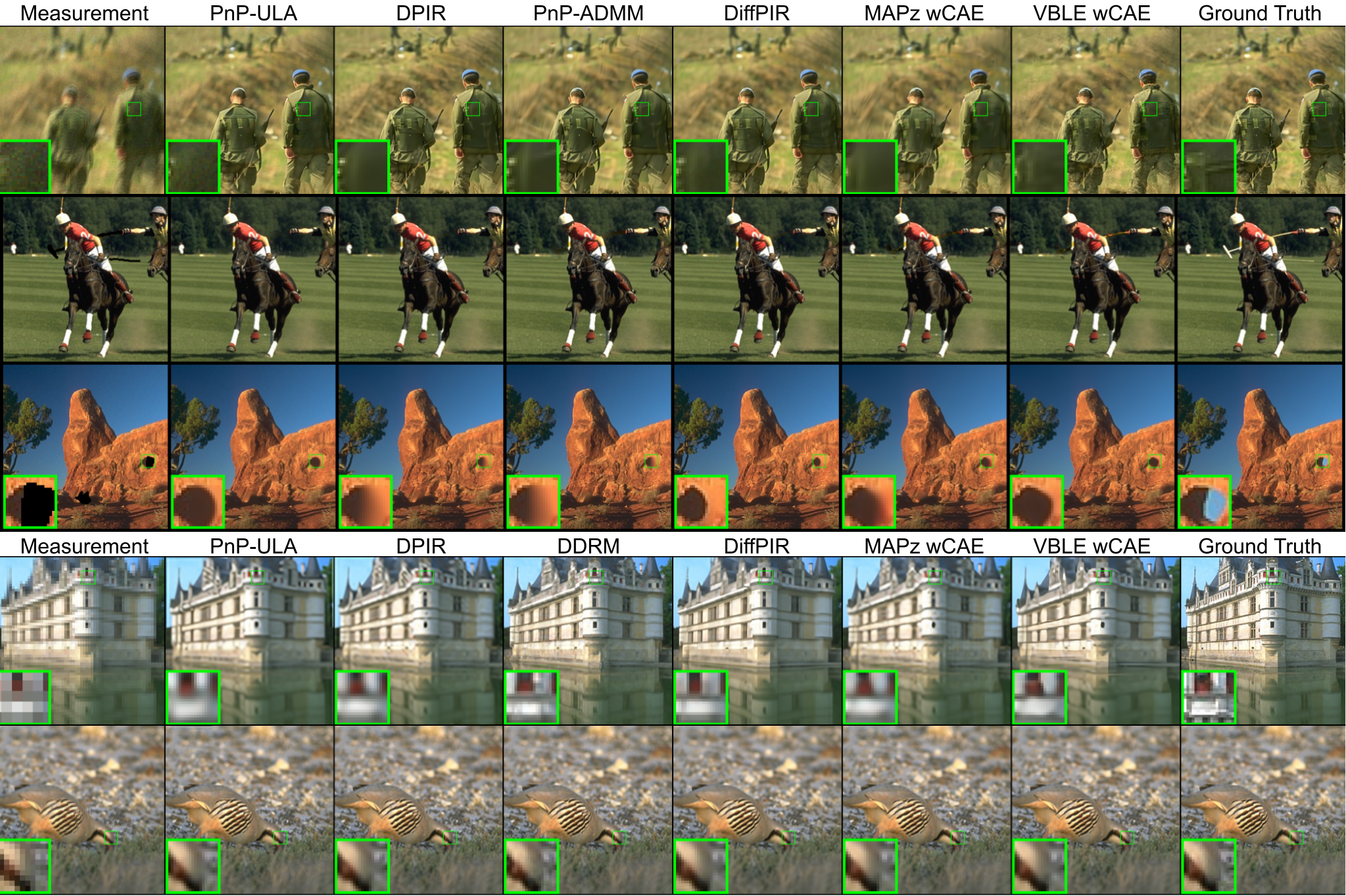}
    }
    \caption{\footnotesize{Qualitative image restoration results on BSD. From top to bottom: Motion deblurring ($\sigma =7.65/255$), SISR $\times 4$, SISR $\times 2$ and two block inpainting experiments ($\sigma=2.55/255$).}}
    \label{fig:res_bsd}
\end{figure*}
\begin{table}[h]
\centering
\scriptsize
\toadd{
\setlength{\tabcolsep}{1.5pt}
\begin{tabular}{lccccccc}
\hline
 & \multicolumn{2}{c}{\textbf{1 solution}} & \multicolumn{2}{c}{\textbf{100 post.samples}} & \multicolumn{1}{l}{\textbf{GPU mem.}} & \multicolumn{1}{l}{\textbf{\#NN}} & \multicolumn{1}{l}{\textbf{\#Backward}} \\
 & Time & \#NFEs & Time & \#NFEs & (MiB) & \textbf{params} & \textbf{} \\ \hline
VBLE wCAE & 22.4s & 1000 & \textbf{24.3s} & \textbf{1 100} & 400 & \textbf{29M} & 1000 \\ \hdashline
PnP-ULA & x & x & 1h23min & 100 000 & 341 & 32M & 0 \\
DPIR & \textbf{1.9s} & \textbf{50} & x & x & \textbf{280} & 32M & 0 \\
PnP-ADMM & 4.1s & 150 & x & x & 282 & 32M & 0 \\
DiffPIR & 21s & 100 & 35min & 10 000 & 2690 & 552M & 0 \\
DDRM & 44.5s & 100 & 1h14min & 10 000 & 36 270 & 552M & 0 \\ \hline
\label{tab:nparams}
\end{tabular}
}
\caption{\toadd{Complexity analysis of our method and the baselines of Section IV.C-D. \# means "number of". The times, the number of Neural Function Evaluations (NFEs), the number of backward steps, and maximum GPU memory are given for inpainting a BSD image on a Nvidia Quadro RTX 8000. 
Time and \#NFEs are given for computing either 1 estimate or 100 posterior samples (for VBLE, optimization AND sampling included).
} 
}
\label{tab:complexity}
\end{table}

\noindent We now compare VBLE using CAEs to state-of-the-art PnP baselines on BSD dataset, which is composed of various natural images. We consider 
\toadd{four} state-of-the-art methods: DPIR \cite{Zhang2021}, PnP-ADMM \cite{Venkatakrishnan2013}, DiffPIR \cite{Zhu2023} \toadd{and DDRM \cite{Kawar2022}. As for FFHQ, we do not perform motion blur with DDRM as it is not implemented in the source code.}  
We also compare to the MCMC approach PnP-ULA \cite{Laumont2022a}, which is a state-of-the-art posterior sampling method. 
For DPIR, PnP-ADMM, and PnP-ULA, we use an available version of DRUNet \cite{Zhang2021} trained on BSD. For DiffPIR \toadd{and DDRM}, a pretrained model on ImageNet \cite{Dhariwal2021} is used, that we finetune on BSD dataset. For PnP-ULA, $10^5$ MCMC iterations are performed.
Similarly to FFHQ results, we provide restoration results with VBLE and its deterministic counterpart using CAE (denoted by MAPz wCAE in the experiments).

\Cref{tab:res_bsd} shows quantitative results on deblurring and SISR inverse problems, while \cref{fig:res_bsd} provides a visual comparison. VBLE exhibits very consistent performance as it reaches the first or second rank for a majority of inverse problems and metrics. Besides, VBLE clearly outperforms PnP-ULA. Visually, except for PnP-ULA, all methods perform very well. In particular, \toadd{the diffusion-based results are particularly sharp, while} VBLE shows very relevant results, staying faithful to the image. 


Additionally, 
\toadd{\cref{tab:complexity}} provides 
\toadd{insights about the computation time, maximum GPU memory, and other complexity-related information. VBLE time includes the variational inference optimization and the sampling for one restoration. In particular, for drawing 100 posterior samples, VBLE is much faster than DiffPIR, DDRM, and PnP-ULA, with a lower computational load than DiffPIR and DDRM. \toaddminor{Note that \cref{alg_ir} is performed during 1000 iterations, thus VBLE, even for computing a single estimate, requires 1000 NFEs.}}

Therefore, VBLE yields state-of-the-art image restoration results compared to very competitive diffusion and \toadd{PnP }
baselines while outperforming the posterior sampling methods in terms of computation time and GPU load. This demonstrates the huge interest of VBLE with CAE for image restoration tasks, especially when uncertainty quantification is required as it enables posterior sampling with a \toadd{far lower }
computational cost \toadd{than} 
MCMC or diffusion\toadd{-}based methods.

\subsection{Posterior distribution quality assessment}
\label{sec:uncertainty}

\begin{table*}[ht]
\scriptsize
\centering
\setlength{\tabcolsep}{4pt}
\begin{tabular}{llcccccccccccccc}
    \hline
    \textbf{FFHQ} & & \multicolumn{3}{c}{\textbf{Deblur} (Gaus.) $\sigma_k=1$} & \multicolumn{3}{c}{\textbf{Deblur} (Motion)} & \multicolumn{3}{c}{\textbf{SISR} $\times 2$} & \multicolumn{3}{c}{\textbf{SISR} $\times 4$} \\ \cline{2-14} 
    \textbf{Method} & Time & PSNR $\uparrow$ & LPIPS $\downarrow$ & ICP 95\% & PSNR $\uparrow$ & LPIPS $\downarrow$ & ICP 95\% & PSNR $\uparrow$ & LPIPS $\downarrow$ & ICP 95\% & PSNR $\uparrow$ & LPIPS $\downarrow$ & ICP 95\% \\
    \hline
    VBLE wCAE & \textbf{27sec} & \underline{34.48} & \underline{0.1473} & \toadd{\underline{0.85}} & \underline{32.22} & \underline{0.1933} & \toadd{\underline{0.87}} & \textbf{36.16} & \underline{0.1251} & \toadd{0.57} & \underline{31.18} & \underline{0.2290} & \toadd{\textbf{0.73}} \\
    DiffPIR-100 & 8min10sec & \textbf{34.78} & \textbf{0.1445} & 0.76 & \textbf{32.95} & \textbf{0.1869} & 0.75 & \underline{35.78} & \textbf{0.1207} & \underline{0.79} & \textbf{31.21} & \textbf{0.2252} & \underline{0.69} \\
    PnP-ULA & 1h23min & 32.51 & 0.2014 & \textbf{0.99} & 30.35 & 0.2367 & \textbf{0.99} & 30.82 & 0.1570 & \textbf{0.92} & 28.94 & 0.2495 & 0.68 \\
    \hline
\end{tabular}
\centering
\begin{tabular}{llcccccccccccccc}
    \hline
    \textbf{BSD} & & \multicolumn{3}{c}{\textbf{Deblur} (Gaus.), $\sigma_k=1$} & \multicolumn{3}{c}{\textbf{Deblur} (Motion)} & \multicolumn{3}{c}{\textbf{SISR} $\times 2$} & \multicolumn{3}{c}{\textbf{SISR} $\times 4$} \\ \cline{2-14} 
    \textbf{Method} & Time & PSNR $\uparrow$ & LPIPS $\downarrow$ & ICP 95\% & PSNR $\uparrow$ & LPIPS $\downarrow$ & ICP 95\% & PSNR $\uparrow$ & LPIPS $\downarrow$ & ICP 95\% & PSNR $\uparrow$ & LPIPS $\downarrow$ & ICP 95\% \\
    \hline
    VBLE wCAE & \textbf{27sec} & \underline{29.62} & \underline{0.1901 }& \toadd{\underline{0.81}} & \underline{28.38} & \underline{0.2356} & \toadd{\underline{0.78}} & \underline{29.55} & 0.1802 & \toadd{0.44} & \textbf{25.53} & \underline{0.3507} & \toadd{\underline{0.60}} \\
    DiffPIR-100 & 35min00sec & \textbf{29.88} & \textbf{0.1847} & 0.67 & \textbf{29.09} & \textbf{0.2351} & 0.70 & \textbf{29.61} & \underline{0.1789} & \underline{0.61} & \underline{25.36} & \textbf{0.3394} & \textbf{0.69} \\
    PnP-ULA & 1h23min & 27.71 & 0.2675 & \textbf{0.95} & 27.17 & 0.2643 & \textbf{0.98} & 28.42 & \textbf{0.1659} & \textbf{0.78} & 24.82 & 0.3730 & 0.59 \\
    \hline
\end{tabular}
\caption{\toadd{Comparison of posterior sampling methods. DiffPIR-100 corresponds DiffPIR, which is run 100 times for each image to get 100 posterior samples. The time is the computation time needed for drawing 100 posterior samples for an inpainting problem on a Nvidia Quadro RTX 8000 (for VBLE time, optimization AND sampling is included). ICP 95\% is the confidence coverage probability of level 95\% (the closer to 0.95, the better). $\sigma=7.65$ for deblurring.}}
\label{tab:ps_methods}
\end{table*}

\begin{figure*}[ht!]
    \centering
    \addbox{
    \includegraphics[width = 0.95\textwidth]{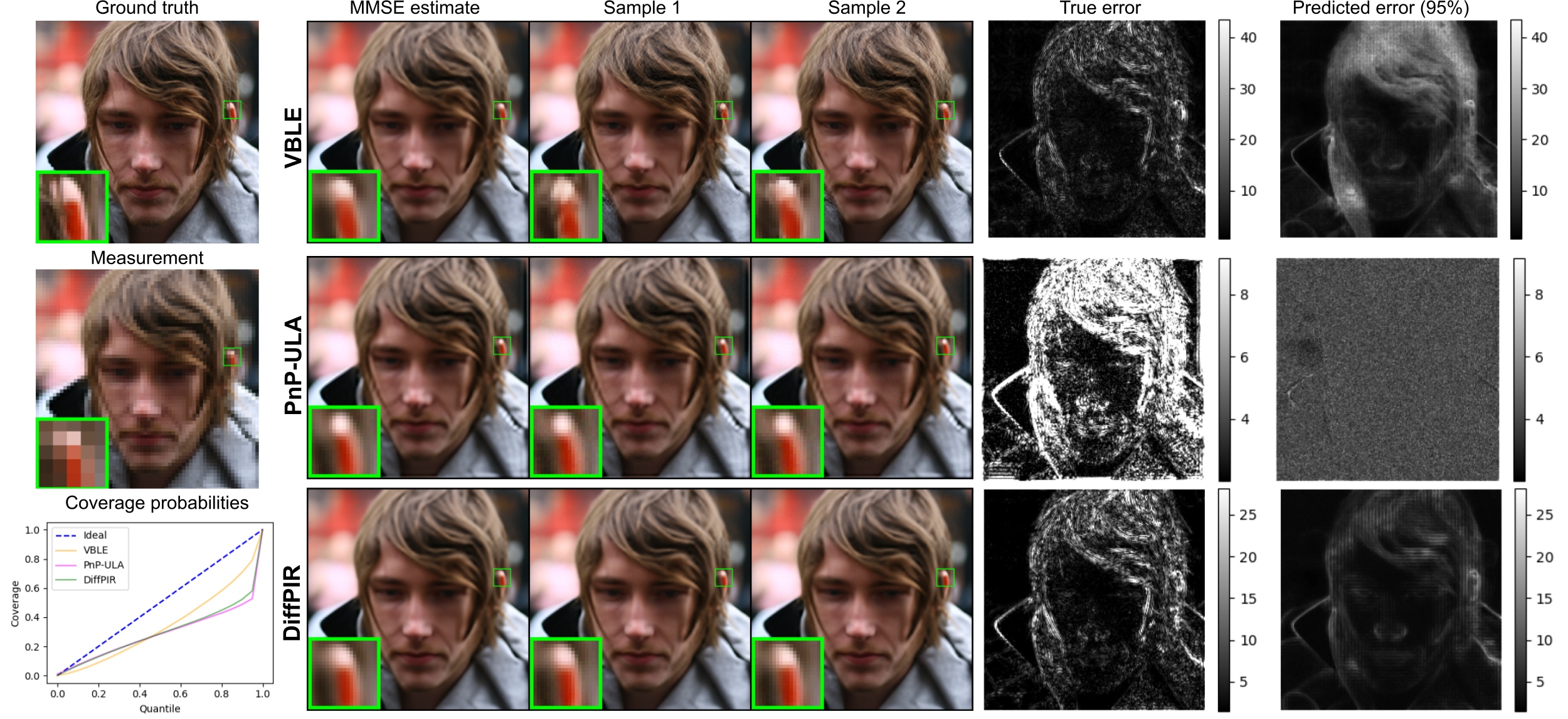}
    }

    \addbox{
    \includegraphics[width = 0.95\textwidth]{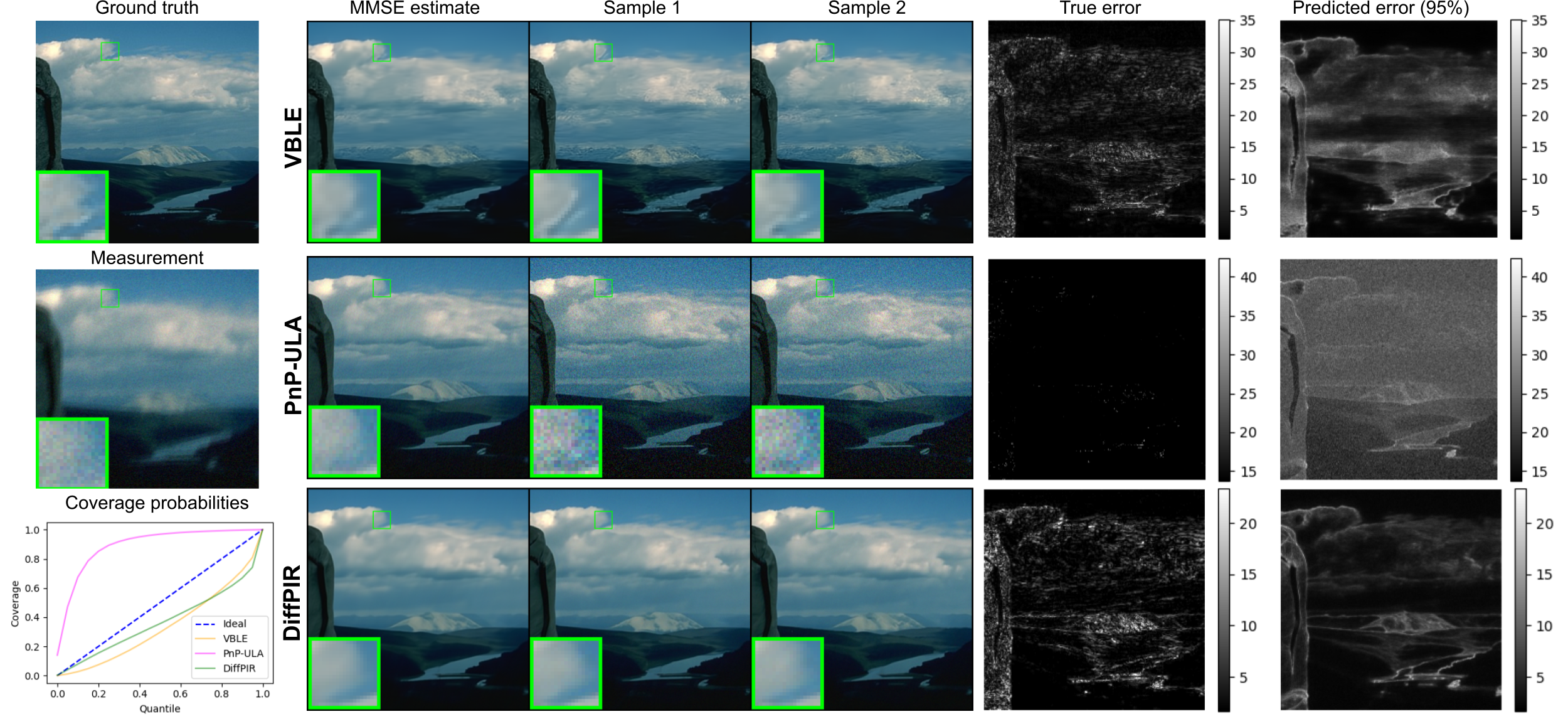}
    }
    \caption{\footnotesize{Visual comparison of posterior sampling ability of VBLE, DiffPIR and PnP-ULA. Top: SISR $\times 2$ on FFHQ. Bottom: Motion deblurring on BSD. The MMSE estimate is the average of 100 samples for each method. The true error denotes the absolute error $|x - \hat{x}|$ between the ground truth and the MMSE. The predicted error (95\%) denotes the 95\% quantile of $\{ |\hat x - x_i| \}_i$} with $x_i$ a posterior sample.}
    \label{fig:res_uncert}
\end{figure*}
\begin{figure*}[ht!]
    \centering
    \addboxminor{
    \includegraphics[width = 1\textwidth]{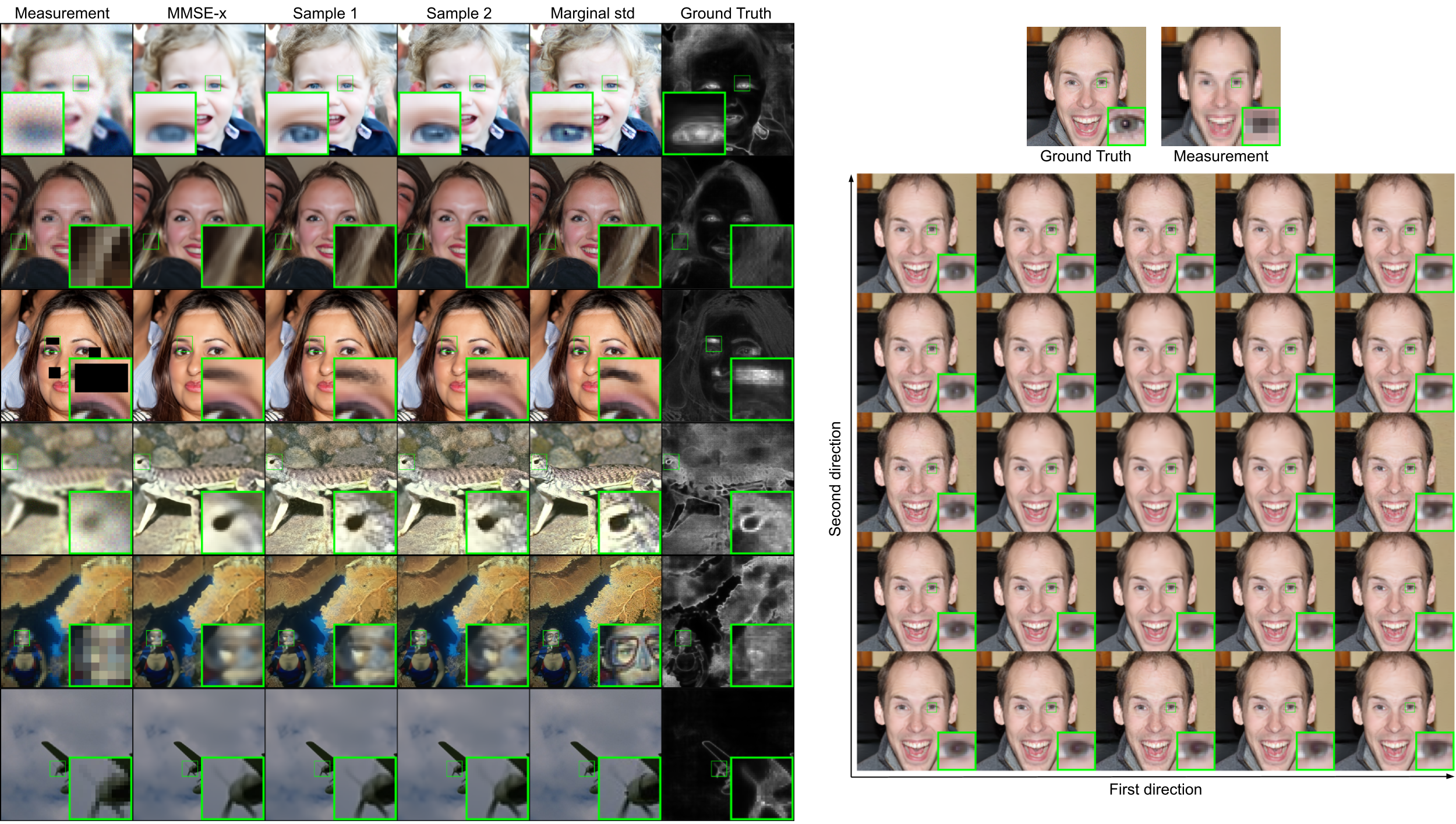}
    }
    \caption{\footnotesize{\toadd{Example of VBLE posterior sampling ability. \toaddminor{Left:} The MMSE-x, which is the estimate used in all quantitative comparison, corresponds to the average over samples in the image space. Marginal std corresponds to pixelwise marginal standard deviations computed with 100 posterior samples. \toaddminor{Right: Illustration of the latent diversity on FFHQ for a SISR $\times 4$ problem. The samples lie on a grid. The coordinates correspond to the latent coordinates of the samples, projected on two orthogonal directions with high variance in the latent space.}}}}
    \label{fig:res_sampling}
\end{figure*}

\noindent Sections \ref{sec:ffhq} and \ref{sec:bsd} show that VBLE using CAEs reaches state-of-the-art point estimation results for image restoration while outperforming the other posterior sampling methods. In this section, we analyze the quality of VBLE approximate posterior distribution, qualitatively and quantitatively \toadd{compared to DiffPIR and PnP-ULA}. 
First, \toadd{\cref{tab:ps_methods} provides an in\toadd{-}depth comparison between VBLE, DiffPIR, and PnP-ULA. In this table, the metrics are computed from the posterior mean, estimated by the average of 100 posterior samples for each image. Thus, DiffPIR is run 100 times for each of these images. Finally, the different metrics are averaged on the whole test set. 
DiffPIR has very good PSNR and LPIPS but is far slower than VBLE. Furthermore, in this table, }
we evaluate the quality of the different posteriors using confidence intervals. Precisely, for each pixel of an image, we compute a confidence interval of a given level using 100 posterior samples. Then, we evaluate the interval coverage probability (ICP), that is the proportion of ground truth pixels lying in the predicted interval. In 
\toadd{\cref{tab:ps_methods}}, ICPs at level $95\%$ are provided for several inverse problems. 
\toadd{PnP-ULA has the best ICPs, while VBLE ICPs are slightly below and DiffPIR is outperformed by PnP-ULA and VBLE. Hence, VBLE offers an interesting trade-off as 
it remains competitive with DiffPIR, while having a relevant approximate posterior and above all, being significantly faster than both methods for posterior sampling.}

Second,  in \cref{fig:res_uncert}, we provide, for the different methods, some posterior samples, as well as a comparison between the true pixelwise error and the 95\% quantile of the predicted error. 
A coverage graph is also provided for each image. Precisely, for each image, a confidence interval of level $\beta$ is derived from the squared error for each pixel. Then, the empirical coverage is plotted as a function of the theoretical coverage $\beta$ varying from 0 to 100\%. This evaluation process is the pixelwise counterpart of the one proposed in~\cite{Pereyra2024}.
For both images of \cref{fig:res_uncert}, VBLE exhibits \toadd{on} average better coverage graphs than DiffPIR and PnP-ULA. Visually, the predicted error\toadd{s} are relevant. In comparison, PnP-ULA predictions are over or underestimated depending on the problems, probably due to the difficulty of the MCMC algorithm to explore efficiently the posterior distribution. DiffPIR predictions seem relevant but underestimated in the SISR problem.

Lastly, VBLE posterior sampling ability is shown in more detail in \cref{fig:res_sampling}. \toaddminor{In the left part of the figure,} posterior samples are sharp and realistic, showing the interest of formulating the posterior distribution in the latent space, using a uniform distribution. The MMSE-x estimate is smoother than the samples and, hence, tends to be more faithful to the target image. \toaddminor{Additionally, the right part of \cref{fig:res_sampling} provides an insight into the obtained latent diversity, using posterior samples projected on two orthogonal directions in the latent space with a high variance, that is lying in the set $\{ a \odot u | u \in \{-0.5,0.5\}^{C\times M\times N} \}$ with $a$ the optimized variational parameter from \cref{eq:paramfam}. Following the first direction, the catchlight, that is the white light reflect in the eye, goes from right to left. Following the second direction, its size increases. This illustrates the uncertainty that can be learnt within this latent uncertainty model.}

Therefore, in addition to state-of-the-art point estimation results, VBLE approximate posterior distribution is relevant, as it provides accurate confidence intervals and realistic posterior samples. This validates the choice of the parametric family chosen to estimate the posterior in the latent space of CAEs. In particular, a\toadd{n} unimodal latent distribution with a similar shape as the encoder posterior $q_\phi(z|x)$ shape provides a remarkably simple way to model the posterior for an image restoration task.

\subsection{Ablation Study}
\label{sec:ablation}

\begin{table}[h]
    \centering
\footnotesize
\setlength{\tabcolsep}{1.5pt}
\begin{tabular}{lcccccc}
\hline
\textbf{} & \multicolumn{3}{c}{\textbf{FFHQ - SISR $\times 4$}} & \multicolumn{3}{c}{\textbf{BSD - Gaussian deblur}} \\ \cline{2-7} 
\textbf{Method} & PSNR $\uparrow$ & LPIPS $\downarrow$ & SSIM $\uparrow$ & PSNR $\uparrow$ & LPIPS $\downarrow$ & SSIM $\uparrow$ \\ \hline
VBLE MMSE-x & \toadd{\textbf{31.18}} & \toadd{0.2290} & \toadd{0.8669} & \toadd{24.66} & \toadd{0.3956} & \toadd{0.6433} \\ \hline
VBLE MMSE-z & \toadd{31.17} & \toadd{0.2468} & \toadd{0.8662} & \toadd{\textbf{24.73}} & \toadd{0.4278} & \toadd{\textbf{0.6519}} \\
VBLE$_{z,h}$ MMSE-x & \toadd{\textbf{31.18}} & \toadd{\textbf{0.2281}} & \toadd{\textbf{0.8671}} & \toadd{24.68} & \toadd{\textbf{0.3952}} & \toadd{0.6444}  \\
\hline
\end{tabular}
\caption{Quantitative results for different variants of our method. MMSE-x and MMSE-z  
denote the two possible estimators defined in \cref{sec:vble}. For Gaussian deblur, the noise level is $\sigma=7.65/255$ and kernel deviation is $\sigma_k = 3$.}
\label{variants_tab}
\end{table}

\noindent For the experiments presented in the paper, some choices have been made. In particular, the MMSE-x estimate has been preferred to the MMSE-z estimate (see \cref{sec:vble}), while only $z$ has been optimized as we approximated the second latent variable $h = E_{h,\phi}(\bar z)$ (see \cref{sec:caereg}). 
In this section, in order to justify these choices, we perform quantitative and qualitative results with the MMSE-z estimate and with an optimization on $(z,h)$ (denoted by VBLE$_{z,h}$). 
All these compared variants of the proposed method share the same restoration parameters (\textit{i.e.} same $\lambda$ and $\alpha$) and use the same CAE.

\begin{figure}[ht]
    \centering
    \includegraphics[width = 0.5\textwidth]{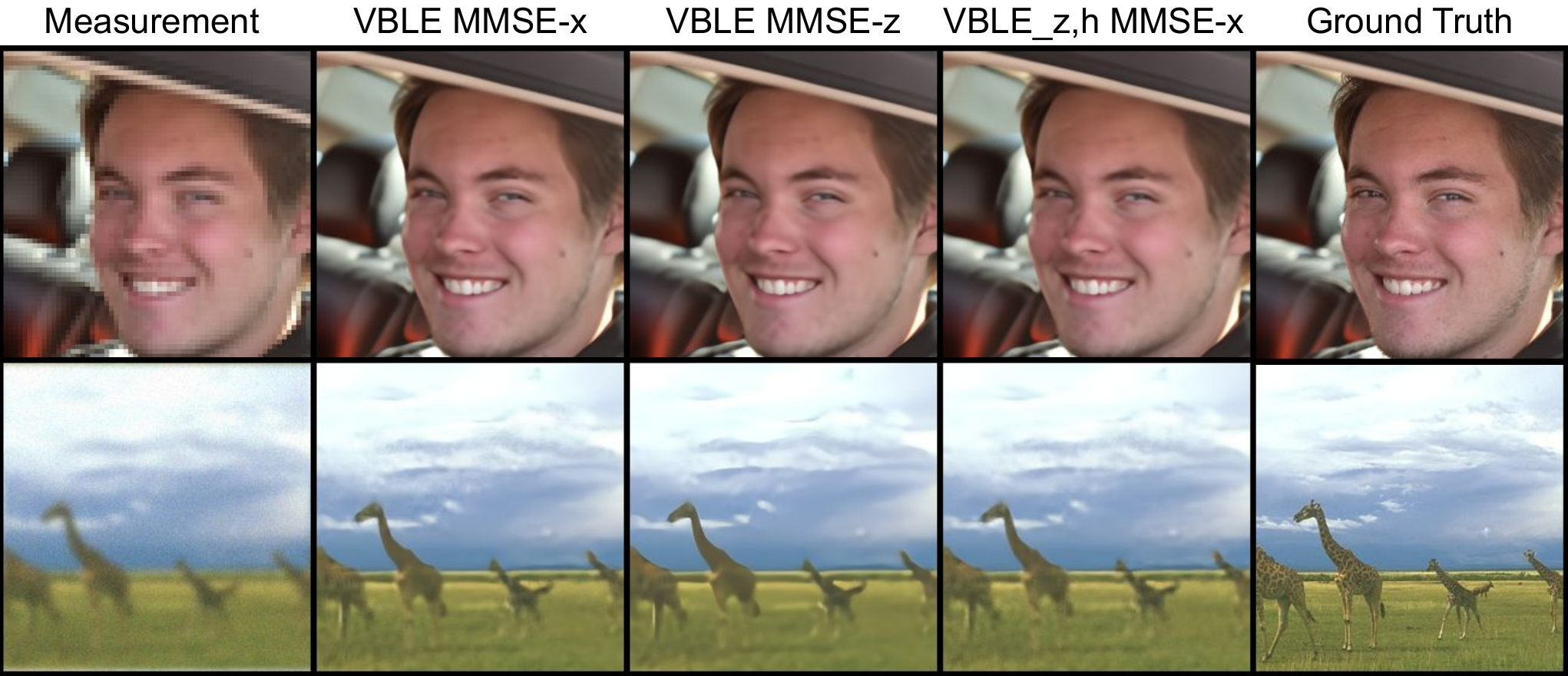}
    \caption{Visual results for different variants of our method. Top: SISR $\times 4$. Bottom: Gaussian deblurring ($\sigma_k = 3$, $\sigma=7.65/255$). }
    \label{variants_visu_fig}
\end{figure}

\Cref{variants_tab} shows quantitative results for VBLE and its variant on two inverse problems.
 VBLE$_{z,h}$ and the reference version yield almost identical results. Thus, the optimization on $(z,h)$ does not perform better than the optimization on $z$ only. For that reason, we choose to optimize only on $z$ for the main experiments, as the formulation is much simpler.
Furthermore, the MMSE-z estimate yields similar PSNR and SSIM results as the MMSE-x estimate but exhibits always a poorer LPIPS. Furthermore, visual results are given in \cref{variants_visu_fig}. While VBLE$_{z,h}$ and the reference version are visually very similar, the MMSE-z estimate is less textured and detailed compared to the MMSE-x, especially on the deblurring problem. This, combined with slightly lower metrics, justifies the use of the MMSE-x estimate for the main experiments.

\begin{figure}[ht]
   \begin{minipage}{0.22\textwidth}
     \centering
     \includegraphics[width=1\linewidth]{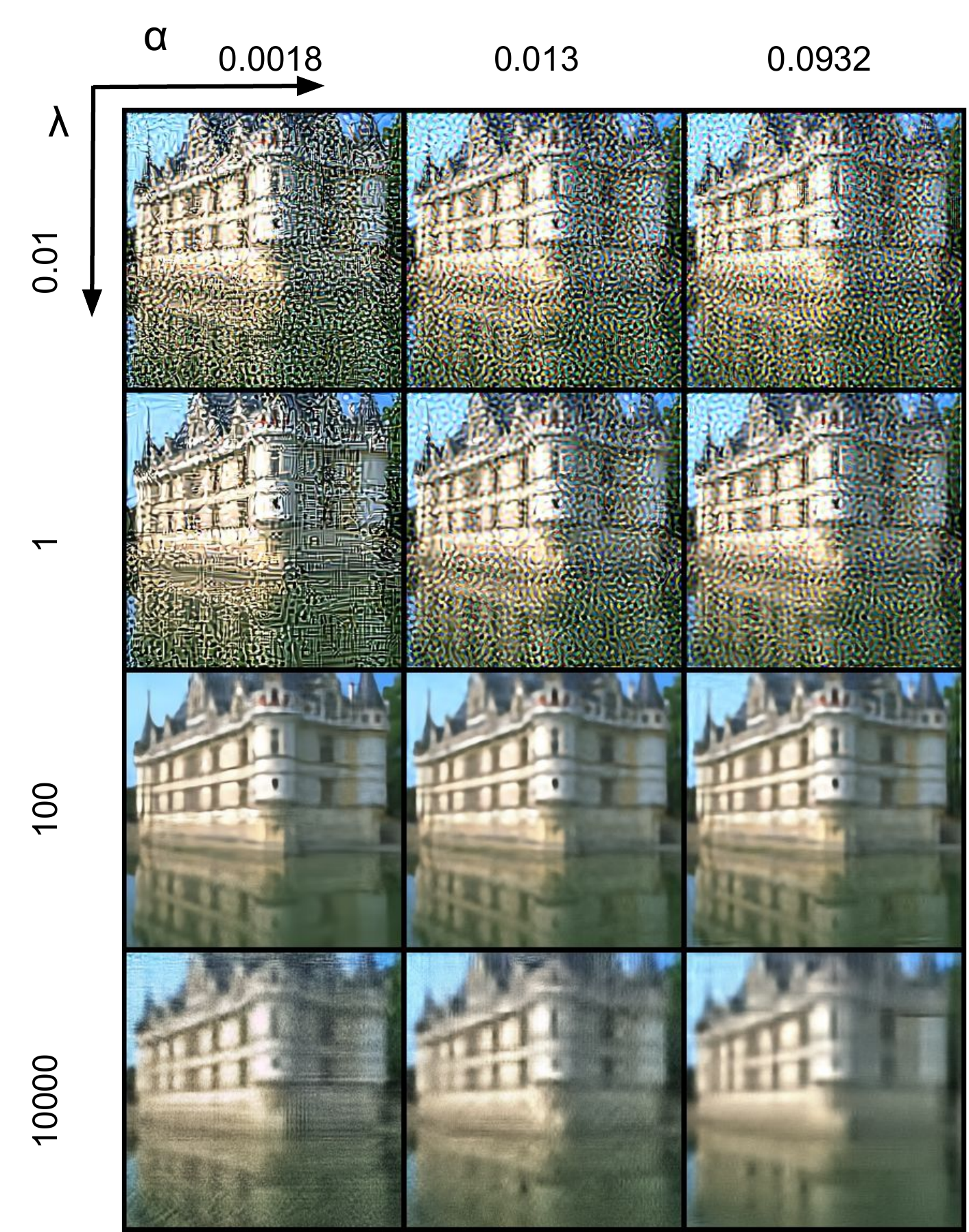}
   \end{minipage} 
   \begin{minipage}{0.22\textwidth}
     \centering
     \includegraphics[width=1\linewidth]{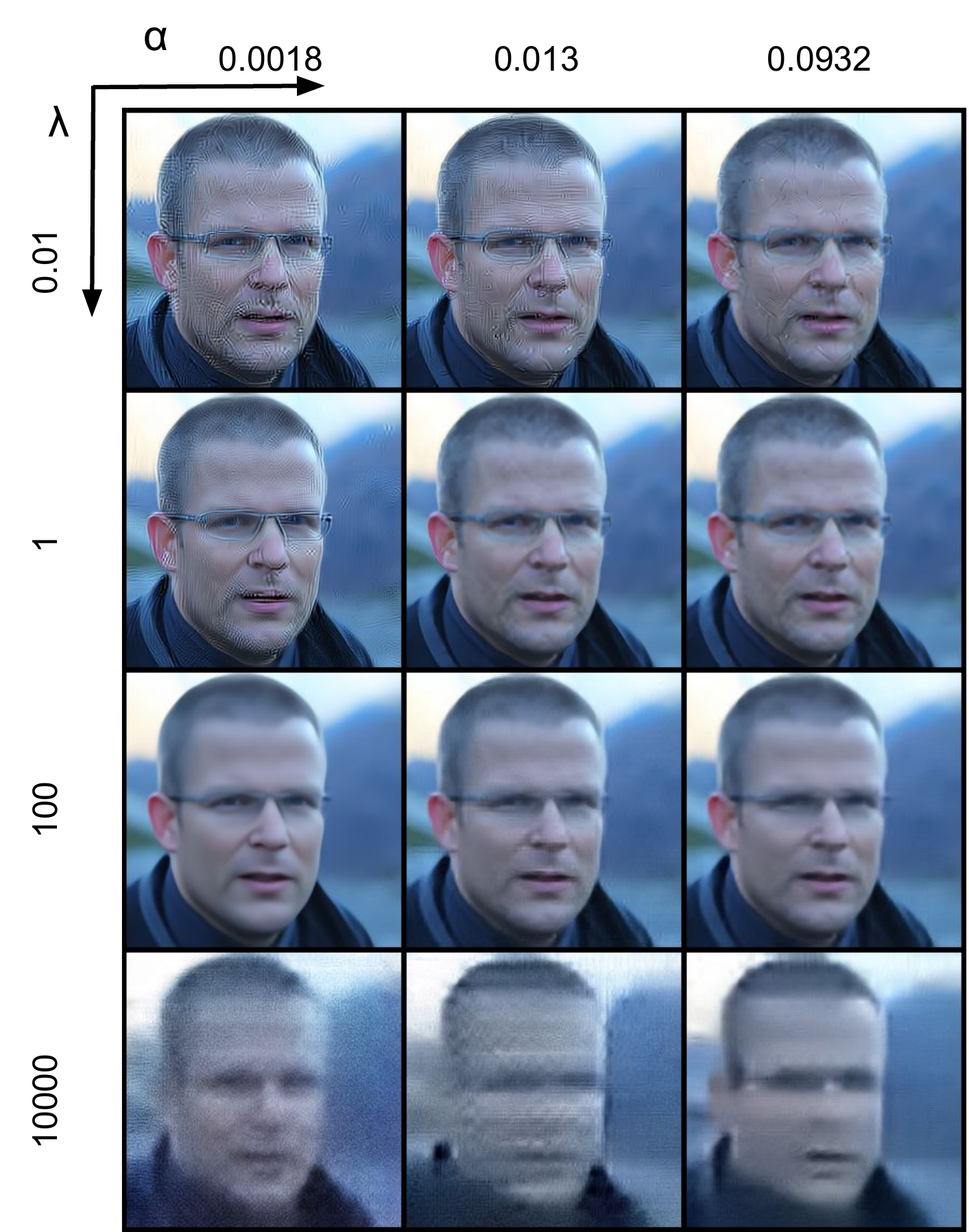}
   \end{minipage}
   \caption{
   Influence of hyperparameters $\lambda$ (regularization parameter) and $\alpha$ (bitrate parameter) on VBLE algorithm. To the left, on a deblurring problem on a BSD image, to the right, on a SISR problem on a FFHQ image.}
   \label{fig_grid_params}
\end{figure}

Additionally, a study on VBLE hyperparameters $\alpha$ (bitrate parameter) and $\lambda$ (regularization parameter) is provided in \cref{fig_grid_params}. Parameter $\lambda$ seems a more sensitive parameter than $\alpha$, hence a CAE with a single bitrate could solve a wide range of inverse problem\toadd{s}. Furthermore, the prior on FFHQ seems to be more informative compared to the one on BSD, which makes sense as BSD dataset structure is much harder to learn.

\section{Conclusion}
In this paper, we have proposed the use of variational compressive autoencoders for regularizing imaging inverse problems. Additionally, we have introduced the VBLE algorithm which performs variational inference in the latent space of CAEs or VAEs.
VBLE enables efficient approximation of the posterior distribution associated with image restoration tasks. 
In our experiments, compressive autoencoders, combined with VBLE, achieve state-of-the-art results and provide compelling posterior sampling abilities, outperforming other posterior sampling methods in terms of computational time and GPU load. These demonstrate the interest of CAEs for image restoration as well as the possibility 
\toadd{of estimating} simply and efficiently the posterior distribution in the latent space of a generative model.

Further work will be dedicated to enhancing the performance of VBLE, by considering more expressive variational distribution families or by a joint estimation in the latent and image space \cite{Gonzalez2022,Duff2022}.
Another key perspective is to consider modulated CAEs~\cite{Song2021} which allow for compression at multiple bit-rates, thus enabling the regularization of inverse problems of various difficulties with a single neural network.

\section*{Acknowledgments}
This work was partly supported by CNES under project name DEEPREG, and ANITI under grant agreement ANR-19-PI3A-0004.

\FloatBarrier
{
    \small
    \bibliographystyle{abbrv}
    \bibliography{biblio}
}

\clearpage
\onecolumn
\appendix

The following \cref{tab_add_exp_setting} and \cref{tab_param_setting} list all the parameters used in the experiments.

\begin{table*}[ht]
\centering
\caption{Detailed experimental settings for VBLE and the baselines. }
\footnotesize{
\begin{tabular}{lll}
\hline
\textbf{Method} & \textbf{Optimized parameters} & \textbf{Fixed parameters} \\ \hline
\textbf{VBLE wCAE} & \begin{tabular}[c]{@{}l@{}}- $\lambda$ (regularization)\\ - $\alpha$ (bitrate)\end{tabular} & - 1000 iterations \\ \hline
\textbf{MAPz wCAE} & \begin{tabular}[c]{@{}l@{}}- $\lambda$ (regularization)\\ - $\alpha$ (bitrate)\end{tabular} & - 1000 iterations \\ \hline
\textbf{DPIR} & \begin{tabular}[c]{@{}l@{}}- $\lambda$ (regularization)\\ - number of iterations\end{tabular} & x \\ \hline
\textbf{PnP-ADMM} & \begin{tabular}[c]{@{}l@{}}- $\lambda$ (regularization)\\ - number of iterations\\ - denoiser noise\end{tabular} & x \\ \hline
\textbf{PnP-ULA} & \begin{tabular}[c]{@{}l@{}}- $\lambda$ (regularization)\\ - denoiser noise\\ - step size of the Markov Chain\end{tabular} & \begin{tabular}[c]{@{}l@{}}- $10^5$ iterations\\ - Thinning 200\\ - Burn-in 0.8\end{tabular} \\ \hline
\textbf{DiffPIR} & \begin{tabular}[c]{@{}l@{}}- $\lambda$ (regularization)\\ - $\zeta$\end{tabular} & - 100 NFEs \\ \hline
\textbf{DDRM} & x & \begin{tabular}[c]{@{}l@{}}- $\eta = 0.85$ \\ - $\eta_b = 1$ \\ - 100 NFEs\end{tabular} \\ \hline
\textbf{PULSE} & x & - 300 iterations \\ \hline
\textbf{\begin{tabular}[c]{@{}l@{}}MAPz wVAE\\ (Bora et al.)\end{tabular}} & - $\lambda$ (regularization) & - 500 iterations \\ \hline
\end{tabular}
}
\label{tab_add_exp_setting}
\end{table*}


\begin{table*}[ht]
\centering
\caption{Detailed parameter setting for VBLE using cheng and mbt compressive networks, for all experiments in the paper. $\lambda$ is the regularization parameter and $\alpha$ the bitrate parameter. Specifically, each CAE is trained with the loss $\mathcal{L} = 255^2 \alpha \mbox{Distortion} + \mbox{Rate}$. For mbt and cheng models, two structures, a light and a bigger one, exist \cite{Begaint2020}, which are used respectively for low and high bitrates. For mbt, high bitrate structure is used for $\alpha \geq 0.013$, and low bitrate structure for $\alpha \leq 0.0067$. For cheng, we have chosen the high bitrate structure for high and low bitrates, as it provides the best image restoration results. 
}
    \scriptsize
\begin{tabular}{ccc|c|c|c|c|c|c|c|c}
\hline
\multirow{2}{*}{\textbf{BSD}} & \multirow{2}{*}{} & \multicolumn{3}{c|}{\textbf{Deblur $\sigma=2.55/255$}} & \multicolumn{3}{c|}{\textbf{Deblur $\sigma=7.65/255$}} & \multicolumn{2}{c|}{\textbf{SISR}} & \textbf{Inpainting} \\
 &  & $\sigma_k=1$ & $\sigma_k=3$ & Motion & $\sigma_k=1$ & $\sigma_k=3$ & Motion & $\times 2$ & $\times 4$ & $50 \%$ \\ \cline{3-11} 
\multirow{2}{*}{\textbf{cheng}} & $\alpha$ & 0.0932 & 0.0067 & 0.0483 & 0.0067 & 0.0035 & 0.0067 & 0.025 & 0.0035 & 0.1800 \\
 & $\lambda$ & 0.7 & 1.4 & 0.7 & 0.7 & 0.7 & 0.7 & 3 & 2.5 & 10 \\ \hline
\multirow{2}{*}{\textbf{mbt}} & $\alpha$ & 0.0483 & 0.0067 & 0.025 & 0.0067 & 0.0035 & 0.0067 & 0.025 & 0.0067 & 0.0932 \\
 & $\lambda$ & 0.7 & 0.8 & 0.6 & 0.7 &0.6 & 0.6 & 2.3 & 3.4 & 4.5 \\ \hline
\end{tabular}
\begin{tabular}{ccc|c|c|c|c|c|c|c|c}
\hline
\multirow{2}{*}{\textbf{FFHQ}} & \multirow{2}{*}{} & \multicolumn{3}{c|}{\textbf{Deblur $\sigma=2.55/255$}} & \multicolumn{3}{c|}{\textbf{Deblur $\sigma=7.65/255$}} & \multicolumn{2}{c|}{\textbf{SISR}} & \textbf{Inpainting} \\
 &  & $\sigma_k=1$ & $\sigma_k=3$ & Motion & $\sigma_k=1$ & $\sigma_k=3$ & Motion & $\times 2$ & $\times 4$ & $50 \%$ \\ \cline{3-11} 
\multirow{2}{*}{\textbf{cheng}} & $\alpha$ & 0.1800 & 0.0067 & 0.0932 & 0.025 & 0.0035 & 0.025 & 0.025 & 0.0.013 & 0.1800 \\
 & $\lambda$ & 0.8 & 2.7 & 1.0 & 1.0 & 1.8 & 0.9 & 4.0 & 4.6 & 2.0 \\ \hline
\multirow{2}{*}{\textbf{mbt}} & $\alpha$ & 0.0932 & 0.0067 & 0.0483 & 0.013 & 0.0035 & 0.0067 & 0.0483 & 0.013 & 0.0932 \\
 & $\lambda$ & 1.0 & 0.6 & 0.9 & 1.0 & 0.9 & 0.8 & 3.2 & 1.8 & 2.0 \\ \hline
\end{tabular}
\label{tab_param_setting}
\end{table*}

\end{document}